\documentclass[sigconf]{acmart}

\AtBeginDocument{%
  \providecommand\BibTeX{{%
    \normalfont B\kern-0.5em{\scshape i\kern-0.25em b}\kern-0.8em\TeX}}}
\usepackage{xspace}
\usepackage{amsthm}
\usepackage{enumitem}
\usepackage{booktabs}
\usepackage{multirow}
\usepackage{graphicx}
\usepackage[normalem]{ulem}
\useunder{\uline}{\ul}{}
\usepackage{float}
\usepackage{subcaption}
\usepackage{balance}
\usepackage[toc,page]{appendix}
\newtheorem{theorem}{Theorem}
\newcommand{\modelname}{\textsf{MStein}\xspace}

\copyrightyear{2023} 
\acmYear{2023} 
\setcopyright{acmlicensed}\acmConference[WWW '23]{Proceedings of the ACM Web Conference 2023}{April 30-May 4, 2023}{Austin, TX, USA}
\acmBooktitle{Proceedings of the ACM Web Conference 2023 (WWW '23), April 30-May 4, 2023, Austin, TX, USA}
\acmPrice{15.00}
\acmDOI{10.1145/3543507.3583529}
\acmISBN{978-1-4503-9416-1/23/04}

\begin{document}

\title{Mutual Wasserstein Discrepancy Minimization for Sequential Recommendation}

\author{Ziwei Fan}
\affiliation{%
  \institution{University of Illinois at Chicago}
  \country{USA}
}
\email{zfan20@uic.edu}

\author{Zhiwei Liu}
\affiliation{%
  \institution{Salesforce AI Research}
  \country{USA}
}
\email{zhiweiliu@salesforce.com}

\author{Hao Peng}
\affiliation{%
  \institution{Beihang University}
  \country{China}
}
\email{penghao@act.buaa.edu.cn}
\authornote{Corresponding author}

\author{Philip S. Yu}
\affiliation{%
  \institution{University of Illinois at Chicago}
  \country{USA}
}
\email{psyu@uic.edu}

\begin{abstract}
Self-supervised sequential recommendation significantly improves recommendation performance by maximizing mutual information with well-designed data augmentations. 
However, the mutual information estimation is based on the calculation of Kullback–Leibler divergence with several limitations, including 
the exponential need of the sample size and training instability. 
Also, existing data augmentations are mostly stochastic and can potentially break sequential correlations with random modifications. 
These two issues motivate us to investigate an alternative robust mutual information measurement capable of modeling uncertainty and alleviating KL divergence's limitations.
  
To this end, we propose a novel self-supervised learning framework based on the \textbf{M}utual Wasser\textbf{Stein} discrepancy minimization~(\modelname) for the sequential recommendation. 
We propose the Wasserstein Discrepancy Measurement to measure the mutual information between augmented sequences. 
Wasserstein Discrepancy Measurement builds upon the 2-Wasserstein distance, which is more robust, more efficient in small batch sizes, and able to model the uncertainty of stochastic augmentation processes. 
We also propose a novel contrastive learning loss based on Wasserstein Discrepancy Measurement. 
Extensive experiments on four benchmark datasets demonstrate the effectiveness of \modelname over baselines. 
More quantitative analyses show the robustness against perturbations and training efficiency in batch size. 
Finally, improvements analysis indicates better representations of popular users\slash items with significant uncertainty. 
The source code is in \url{https://github.com/zfan20/MStein}.
\end{abstract}


\maketitle

\section{Introduction}
Recommender systems have been a prevalent and crucial component in several application scenarios~\cite{rendle2010factorizing, lin2020fill, lin2022phish}. 
Among existing personalized recommendations, sequential recommendation~(SR) attracts increasing interest for its scalability and performance. 
SR predicts the next preferred item for each user by modeling the sequential behaviors of the user and capturing item-item transition correlations.

Existing works of SR include Markovian approaches~\cite{rendle2010factorizing, he2016fusing}, convolution-based approaches~\cite{tang2018personalized, yan2019cosrec}, RNN-based approaches~\cite{hidasi2015session, zheng2019gated}, and Transformer-based methods~\cite{kang2018self, sun2019bert4rec, li2020time}. 
The recent success of Self-Supervised Learning~(SSL) further improves SR~\cite{zhou2020s3, xie2022contrastive, liu2021contrastive} by alleviating the data sparsity issue and improving robustness with novel data augmentations and contrastive loss, \textit{i.e.,} InfoNCE. 
As the widely used SSL framework, contrastive learning~(CL) constructs positive and negative pairs via data augmentation strategies. 
The commonly adopted CL loss InfoNCE maximizes the mutual information of positive pairs among all pairs. 
With data augmentations and mutual information maximization, SSL-based SR methods capture more robust user preferences. 

Despite the effectiveness of SSL for SR, we argue that existing SSL for SR methods still suffer critical issues in both data augmentations and the mutual information maximization CL loss due to the following reasons:
\begin{itemize}[leftmargin=*]
    \item \textbf{Stochasticity of Data Augmentations:} Most data augmentation techniques are random augmentations, such as the random sequence crop from CL4Rec and dropout augmentation from DuoRec. Different random augmentations can be viewed as augmentation distributions, and the perturbed sequences are realized samples from augmentation distributions. However, existing CL methods only measure the similarities between realized samples without considering the uncertainty of augmentation distributions. Ignoring uncertain information affects the stability of training and robustness of item embeddings learning against noises. 
    \item \textbf{Limitations of KL-divergence-based Mutual Information Measurement:} Existing methods mainly adopt the InfoNCE as CL loss, which is building upon mutual information maximization. Although DuoRec~\cite{qiu2022contrastive} adopts the alignment and uniformity losses, it still follows \cite{wang2020understanding} to interpret InfoNCE as the combination of alignment and uniformity. We argue that mutual information measurement has several limitations, which originate from KL-divergence, including 
    the exponential need for sample size and instability against small perturbations. All these issues might significantly affect the modeling in CL, especially when augmentations are stochastic and potentially destroy the sequential correlations. 
\end{itemize}

\begin{figure}[]
\centering
\includegraphics[width=0.45\textwidth]{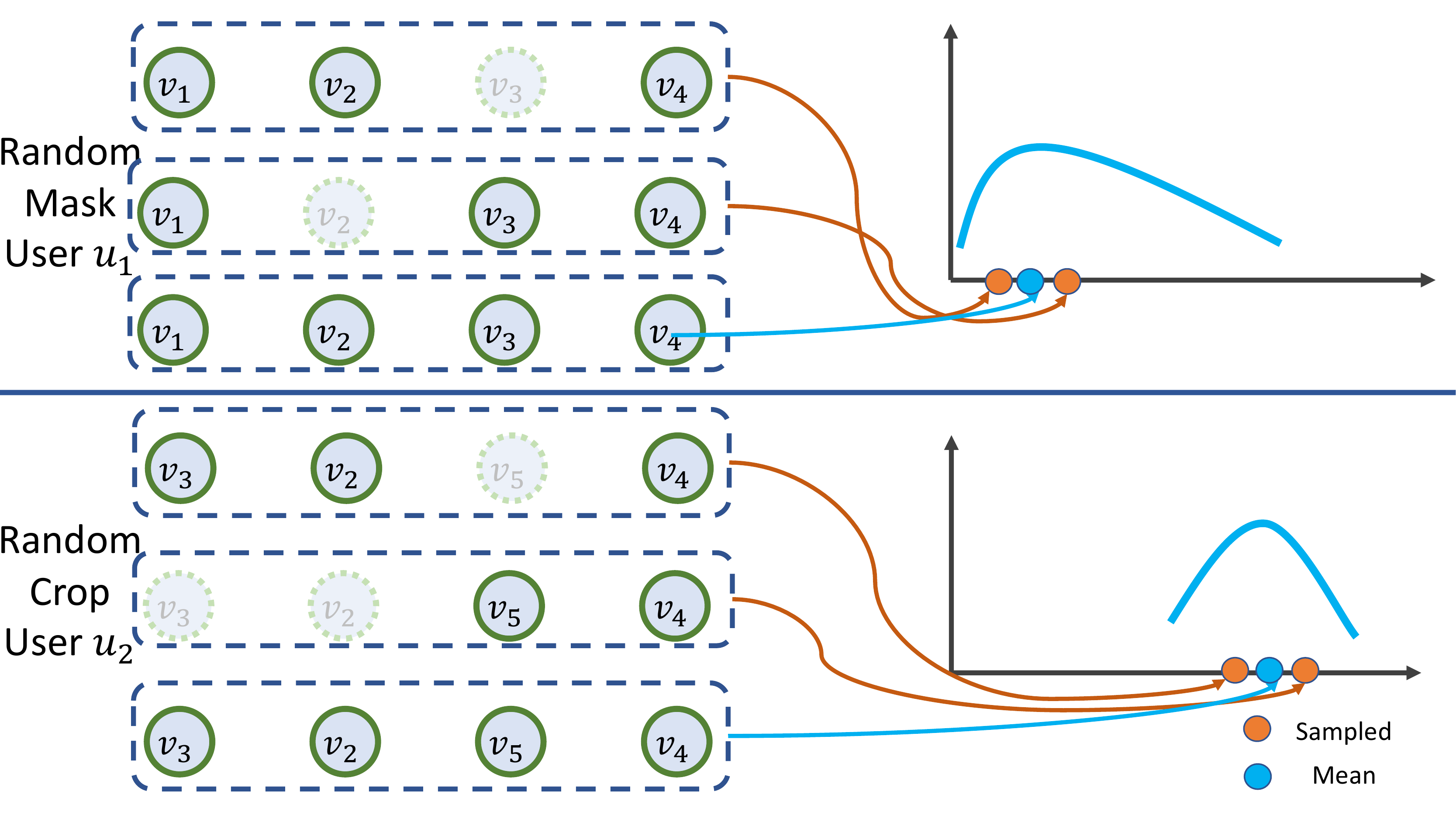}
\caption{Motivation Examples. We only show the random mask and crop as examples but other data augmentations can also be viewed as sampling from augmentation distributions.}
\label{fig:motivaiton}
\end{figure}

As shown in Fig.~(\ref{fig:motivaiton}), each data augmentation strategy has user-specific augmentation distributions. In Fig.~(\ref{fig:motivaiton}), user $u_1$ has a smoother distribution than the user $u_2$. 
Different augmentation strategies introduce different levels of uncertainty and the possibility of breaking the sequential correlations. 
In this example, the random crop is more likely to break the user $u_2$'s sequential correlations. 
Therefore, depending on users, different data augmentations follow augmentation distributions with varying uncertainties, and modeling this stochasticity becomes crucial for learning more robust embeddings. Moreover, distributions of users $u_1$ and $u_2$ have limited overlapping. 
In this case, the calculation of KL-divergence becomes unstable.

These issues motivate us to model the stochasticity of data augmentations and address the limitations of KL-divergence-based mutual information measurement. 
However, developing a framework that achieves these two goals simultaneously is nontrivial because it demands the framework to:
(1) consider uncertainty information in modeling stochasticity data augmentations; 
(2) measure mutual information with uncertainty signals while still bypassing the limitations of KL-divergence.

In this research, we first theoretically analyze the limitations of KL-divergence in mutual information measurement and demonstrate the necessity of proposing the alternative. 
Then, based on the theoretical analysis, we propose the Wasserstein Discrepancy Measurement, which measures mutual information with a 2-Wasserstein distance between distributions. 
Furthermore, we introduce how to adapt Wasserstein Discrepancy Measurement to the contrastive learning framework by proposing the \textbf{M}utual Wasser\textbf{Stein} discrepancy minimization~(\modelname). 
Finally, we build the proposed \modelname with a novel contrastive learning loss based on Wasserstein discrepancy measurement, which greatly advances both STOSA~\cite{fan2022sequential} for stochastic embeddings and CoSeRec~\cite{liu2021contrastive} for data augmentation in sequential recommendation. 

We summarize our four major contributions as follows:
\begin{itemize}[leftmargin=*]
    \item We propose the Wasserstein Discrepancy Measurement as the alternative to existing mutual information measurement based on KL-divergence, which has several theoretically proven limitations as the core component of InfoNCE contrastive loss.
    \item On top of the proposed Wasserstein Discrepancy Measurement, we propose the mutual Wasserstein discrepancy minimization as a novel contrastive learning loss and demonstrate its superiority in modeling augmentation stochasticity and robust representation learning against perturbations.
    \item We show that alignment and uniformity properties are exactly optimized in the proposed Wasserstein Discrepancy Measurement, and the original version of alignment and uniformity optimizations is a version of Wasserstein Discrepancy Measurement.
    \item Extensive experiments first demonstrate the effectiveness of \modelname in generating recommendations over state-of-the-art baselines DuoRec, CL4Rec, and CoSeRec with improvements on all standard metrics from 2.24\% to 20.10\%. Further analysis in robustness and batch sizes show the advantages of adopting Wasserstein Discrepancy Measurement in mutual information measurement.
\end{itemize}

\section{Related Work}
Several topics are related to this paper, including sequential and self-supervised recommendations. 
We first introduce relevant work in the sequential recommendation, which is the problem setting in our paper. 
Then we discuss some related works in the self-supervised recommendation. 
Lastly, we discuss existing works on uncertainty modeling and distinguish our proposed framework from them.

\subsection{Sequential Recommendation}
Sequential Recommendation~(SR)~\cite{peng2020m2, fan2021continuous, 10.1145/3404835.3463036, fan2022sequential} formats each user's temporal interactions as a sequence and encodes the sequential behaviors as the user preference. 
The fundamental idea of SR is modeling item-item transitions within sequences and inferring user preferences from transitions. The earliest work is Markovian approaches, including most notable works FPMC~\cite{rendle2010factorizing} and Fossil~\cite{he2016fusing}. 
Based on the Markov Chain's ability to learn orders of transitions, FPMC learns first-order item-item transitions, assuming that the next item depends on only the previous item. Fossil further combines FPMC and matrix factorization by additional item similarities information. 
With the perspective of viewing the sequence as an image, another line of works based on Convolution Neural Network~(CNN) emerges, such as Caser~\cite{tang2018personalized}, CosRec~\cite{yan2019cosrec}, and NextItNet~\cite{yuan2019simple}. Caser proposes vertical\slash horizontal convolution operations on the sequence embedding. CosRec interprets the sequence as a tensor and adopts the 2d convolution. 
NextItNet utilizes 1D dilated convolution in the sequence embedding. Recurrent Neural Network~(RNN) has shown remarkable performance in the recommendation to utilize sequential information further. 
The relevant works based on RNN include GRU4Rec~\cite{hidasi2015session}, HGN~\cite{ma2019hierarchical, peng2021ham}, HRNN~\cite{quadrana2017personalizing}, and RRN~\cite{wu2017recurrent}. 
These methods adopt different RNN frameworks for SR. HGN and HRNN both propose hierarchical RNNs for SR. With the inspiration of the recent success of the self-attention mechanism, Transformer-based approaches increasingly attract interest in SR.
SASRec~\cite{kang2018self} firstly adopts Transformer in SR. BERT4Rec~\cite{sun2019bert4rec} extends SASRec with bi-directional self-attention module. 
Multiple following works~\cite{fan2021modeling, li2020time, wu2020sse, 10.1145/3404835.3463036, fan2022sequentialmt4sr} build upon SASRec and further enhance SR. 
The advantage of self-attention for SR is its ability to model long-term dependency within the sequence.

\subsection{Self-Supervised Contrastive Learning}
Self-Supervised Learning~(SSL)~\cite{oord2018representation} advances the representation learning in multiple areas, including computer vision~\cite{chen2020simple, he2020momentum, grill2020bootstrap}, NLP~\cite{gao2021simcse, meng2021coco}, graph learning~\cite{you2020graph, zhu2021graph} and recommender systems~\cite{liu2021contrastive, xie2022contrastive, qiu2022contrastive, zhou2020s3, fan2023episodes}. 
Contrastive learning~(CL), as the most widely adopted approach in SSL, improves the model encoder with data augmentations and contrastive loss. 
Data augmentation strategies generate perturbed views of original input data. Then, contrastive learning pulls embeddings of views from the same input closer while pushing away embeddings of views from different inputs. 
Both data augmentations and contrastive loss are crucial components in CL. The most used contrastive loss is InfoNCE~\cite{oord2018representation}, which adopts the categorical cross-entropy loss and aims to maximize mutual information of two variables~(in CL, two perturbed views). \cite{ozair2019wasserstein} proposes a contrastive learning loss based on Euclidean distance to generalize contrastive predictive coding.

Depending on application scenarios, several data augmentation strategies are proposed, and most are stochastic processes for perturbing the original input. 
In computer vision, SimCLR~\cite{chen2020simple} proposes several simple stochastic data augmentations for images in the contrastive learning framework, including distortion, crop, and blurring. 
For recommendation, most existing works in CL for SR propose reorder, mask, and crop as augmentations~\cite{xie2022contrastive}. S3-Rec~\cite{zhou2020s3} utilizes items' attributes and develops a self-supervised pre-training framework for SR. ICLRec~\cite{chen2022intent} applies intent learning into the CL framework. EC4SRec~\cite{wang2022explanation} extends CoSeRec~\cite{liu2021contrastive} with contrastive signals selected by explanation methods. DuoRec~\cite{qiu2022contrastive} proposes supervised contrastive signals and dropout as unsupervised signals in CL for SR.
Existing works commonly propose augmentation methods specific to the application, and most augmentations are stochastic.

\subsection{Uncertainty Modeling}
Modeling uncertainty information has been attracting interest from the research community~\cite{vilnis2014word, bojchevski2018deep, sun2018gaussian, he2015learning}. 
The uncertainty information is categorized into aleatoric and epistemic uncertainty~\cite{hullermeier2019aleatoric}. 
Aleatoric uncertainty describes the stochastic uncertainty that is unavailable to be known, while epistemic uncertainty describes the systematic uncertainty that is known but hard to measure. 
The most common approach to consider uncertainty information modeling is adopting the Gaussian embedding. For example, DVNE~\cite{zhu2018deep} represents nodes as distributions. 
DDN~\cite{zheng2019deep} and PMLAM~\cite{ma2020probabilistic} interpret users\slash items as distributions for recommendation.
DT4SR~\cite{fan2021modeling} and STOSA~\cite{fan2022sequential} both represent items as Gaussian and develop self-attention adaptive to Gaussian embeddings for SR. 
Moreover, \cite{wang2022rethinking} proposes the AUR framework for modeling interactions' aleatoric uncertainty.

\section{Preliminaries}

\subsection{Problem Definition}
In SR, we have a set of users $\mathcal{U}$ and items $\mathcal{V}$ and the associated interactions.
We can sort interacted items of each user $u\in\mathcal{U}$ based on timestamps in a sequence as $\mathcal{S}^{u}={[v^{u}_1,v^u_2,\dots,v^u_{|\mathcal{S}^{u}|}]}$, where $v^{u}_i\in\mathcal{V}$ is the $i$-th interacted item in the sequence. 
For each user, SR generates a top-N recommendation list of items as the most likely preferred items in the next action.
Formally, we predict the score $p\left( v_{|\mathcal{S}^{u}|+1}^{(u)}=v \left|  \mathcal{S}^{u} \right.\right)$ and rank scores of all items to generate top-N list. 
The core challenge in SR lies in how to model the user's action sequence $\mathcal{S}^{u}$.

\subsection{Stochastic Transformer for SR}
Transformer has been the successful backbone model for SR~\cite{kang2018self,sun2019bert4rec} because of its self-attention module for modeling weights from all items in the sequence. 
However, the original Transformer architecture fails to model uncertainty information in sequences. 
Among existing Transformer variants, STOSA~\cite{fan2022sequential} extends the Transformer to introduce stochastic embeddings and a Wasserstein self-attention module for modeling uncertainty information in SR. 
In this research, we build on STOSA to model uncertainty information. 
Specifically, STOSA represents items as Gaussian distributions with mean and covariance embeddings, which together are defined as stochastic embeddings as follows:
\begin{align}
    \mathbf{E}^{\mu} = \text{Emb}_{\mu}(\mathcal{S}^{u}),\vspace{3mm}\mathbf{E}^{\Sigma} = \text{Emb}_{\Sigma}(\mathcal{S}^{u}).
\end{align}
For the item $v_i$ in $\mathcal{S}^{u}$, its stochastic embeddings proposed by STOSA includes $\mathbf{E}^{\mu}_{v_i}$ and $\mathbf{E}^{\Sigma}_{v_i}$ and parameterizes the Gaussian distribution $\mathcal{N}(\mathbf{E}^{\mu}_{v_i}, \text{diag}(\mathbf{E}^{\Sigma}_{v_i}))$. 
To calculate the self-attention values on a specific pair of items $(v_i,v_j)$, the Wasserstein self-attention adopts the negative 2-Wasserstence distance as follows:
\begin{align}
    \mathbf{A}_{ij} &= -(W_2(v_i, v_j)),
    = -\left(||\mu_{v_i}-\mu_{v_j}||^2_2 + ||\Sigma_{v_i}^{1/2}-\Sigma_{v_j}^{1/2}||^2_{\text{F}}\right),
\end{align}
where $W_2(\cdot, \cdot)$ denotes the 2-Wasserstence distance, $\mu_{v_i} = \mathbf{E}^{\mu}_{v_i}W^{\mu}_K$, $\Sigma_{v_i}=\text{ELU}\left(\text{diag}(\hat{\mathbf{E}}^{\Sigma}_{v_i}W^{\Sigma}_K)\right)+1$, $\Sigma_{v_j}=\text{ELU}\left(\text{diag}(\hat{\mathbf{E}}^{\Sigma}_{v_j}W^{\Sigma}_Q)\right)+1$, $\mu_{v_j} = \mathbf{E}^{\mu}_{v_j}W^{\mu}_Q$, and ELU is the Exponential Linear Unit activation function, $W^{\mu}_K$, $W^{\Sigma}_K$, $W^{\mu}_Q$, and $W^{\Sigma}_Q$ are linear mappings for stochastic embeddings. 
STOSA also has \textit{Feed-forward Neural Networks}, \textit{Residual Connection}, and \textit{Layer Normalization} modules, similar to original Transformer. 
Hence, we formulate the STOSA sequence encoder as:
\begin{align}
    \textbf{h}_{u}=(\textbf{h}_u^{\mu},\textbf{h}_u^{\Sigma}) = \text{StosaEnc}(\mathcal{S}^{u}),
\end{align}
where $\textbf{h}_u^{\mu}$ and $\textbf{h}_u^{\Sigma}$ are stochastic sequence embeddings of $\mathcal{S}^u$, and for each timestep $t$, $\textbf{h}_{u,t}=(\textbf{h}_{u,t}^{\mu}\textbf{h}_{u,t}^{\Sigma})$ encodes the next-item representation. The overall optimization loss is defined as follows:
\begin{align}
\label{eq:rec_loss}
    \mathcal{L}_{\text{rec}} = \sum_{\mathcal{S}^u\in\mathcal{S}}\sum_{t=1}^{|\mathcal{S}^u|}-\log (\sigma(W_2(\textbf{h}_{u,t}, v_t^-) - W_2(\textbf{h}_{u,t}, v_t^+))) + \lambda\ell_{pvn},
\end{align}
where $v_t^+$ is the ground truth next item stochastic embedding, $v_t^-$ denotes the negative sampled item embedding, $\sigma(\cdot)$ denotes the sigmoid activation function, the stochastic embedding tables~($\mu, \Sigma$) are optimized simultaneously, $\ell_{pvn}$ is the positive-vs-negative loss proposed by STOSA. 

\subsection{InfoNCE for Contrastive Learning}
Contrastive loss is the core component of Contrastive Learning ~(CL). 
InfoNCE~\cite{oord2018representation} is the most widely used contrastive loss. 
Minimizing the InfoNCE is equivalent to maximizing the lower bound of mutual information. 
Specifically, given a batch of $N$ user sequences, random data augmentations generate two perturbed views of each sequence, concluding that there are $2N$ sequences, $N$ positive pairs of sequences, and $4N^2-2N$ negative pairs in the InfoNCE calculation. 
We introduce contrastive loss based on the original Transformer encoder. 
For the batch of $N$ user sequences $\mathcal{B}$, the augmentated pairs $S_{\mathcal{B}}$ are:
\begin{align}
    S_{\mathcal{B}}=\{\mathcal{S}^{u_1}_a, \mathcal{S}^{u_1}_b, \mathcal{S}^{u_2}_a, \mathcal{S}^{u_2}_b, \cdots, \mathcal{S}^{u_N}_a, \mathcal{S}^{u_N}_b\},
\end{align}
where subscripts $a$ and $b$ denote two perturbed versions of $\mathcal{S}^{u}$. 
The InfoNCE for a pair of augmented sequences $(\mathcal{S}^{u_i}_a, \mathcal{S}^{u_i}_b)$ is calculated as follows:
\begin{align}
    \mathcal{L}_{cl}(\textbf{h}^{u_i}_a, \textbf{h}^{u_i}_b) = -\log\frac{\exp(\text{sim}(\textbf{h}^{u_i}_a, \textbf{h}^{u_i}_b))}{\exp(\text{sim}(\textbf{h}^{u_i}_a, \textbf{h}^{u_i}_b))+\sum_{j\in S_{\mathcal{B}}^-}\exp(\text{sim}(\textbf{h}^{u_i}_a, \textbf{h}^j))},
\end{align}
where $\textbf{h}^{u_i}_a$ and $\textbf{h}^{u_i}_b$ are sequence embeddings of two perturbed sequences versions learned from the encoder, $S_{\mathcal{B}}^-=S_{\mathcal{B}}-\{\mathcal{S}^{u_i}_a, \mathcal{S}^{u_i}_b\}$ denotes the negative augmented sequence pairs, and the sim$(\cdot)$ denotes the cosine similarity.

\section{Wasserstein Discrepancy Measurement}
In this section, we first recall the definition of mutual information and its connection with InfoNCE in the setting of contrastive learning. 
Then we discuss the disadvantages of KL-divergence-based mutual information measurement. 
Finally, we introduce the proposed Wasserstein Discrepancy Measurement in mutual information measurement to alleviate these disadvantages.

\subsection{InfoNCE and Mutual Information}
InfoNCE in contrastive learning is first adopted in Contrastive Predictive Coding~(CPC)~\cite{oord2018representation}. 
The mutual information is maximized when InfoNCE is optimized. 
Formally, in contrastive learning, we denote the randomly augmented sequences of the user $u_i$ as $(x^{u_i}_a=\mathcal{S}^{u_i}_a, x^{u_i}_b=\mathcal{S}^{u_i}_b)$ and $(x^{u_i}_a, x^{u_i}_b)$ are random variables following random augmentation distributions. 
The connection between InfoNCE and mutual information of $(x^{u_i}_a, x^{u_i}_b)$ is given as:
\begin{align}
\label{eq:infonce_mutual}
    \mathcal{L}_{cl} &= -\mathbb{E}_{\mathcal{S}^{u_i}\in\mathcal{B}}\log\left[\frac{\frac{p(x^{u_i}_a|x^{u_i}_b)}{p(x^{u_i}_b)}}{\frac{p(x^{u_i}_a|x^{u_i}_b)}{p(x^{u_i}_b)}+\sum_{x^-\in S_{\mathcal{B}}^-}\frac{p(x^{u_i}_a|x^-)}{p(x^-)}}\right]\nonumber\\
    &\approx \mathbb{E}_{\mathcal{S}^{u_i}\in\mathcal{B}}\log\left[1+\frac{p(x^{u_i}_b)}{p(x^{u_i}_a|x^{u_i}_b)}(2N-1)\mathbb{E}_{x^-\in S_{\mathcal{B}}^-}\frac{p(x^{u_i}_a|x^-)}{p(x^-)}\right]\nonumber\\
    &=\mathbb{E}_{\mathcal{S}^{u_i}\in\mathcal{B}}\log\left[1+\frac{p(x^{u_i}_b)}{p(x^{u_i}_a|x^{u_i}_b)}(2N-1)\right]\nonumber\\
    &\geq \mathbb{E}_{\mathcal{S}^{u_i}\in\mathcal{B}}\log\left[\frac{p(x^{u_i}_b)}{p(x^{u_i}_a|x^{u_i}_b)}(2N-1)\right]\nonumber\\
    &=-\mathbb{E}_{\mathcal{S}^{u_i}\in\mathcal{B}}\log\left[\frac{p(x^{u_i}_a|x^{u_i}_b)}{p(x^{u_i}_b)}\right]+\log(2N-1)\nonumber\\
    &=\mathbb{E}_{\mathcal{S}^{u_i}\in\mathcal{B}}-I\left(x^{u_i}_a, x^{u_i}_b\right) + \log(2N-1),
\end{align}
where 
$I(x, y)$ is the mutual information between random variables $x$ and $y$. 

Eq.~(\ref{eq:infonce_mutual}) proves that optimizing $\mathcal{L}_{cl}$ simultaneously maximizes mutual information as $I\left(x^{u_i}_a, x^{u_i}_b\right)\geq \log(2N-1)-\mathcal{L}_{cl}$. 
It also shows that when the batch size $N$ grows larger, we can better approximate the mutual information, which has been demonstrated in related works~\cite{liu2021contrastive, xie2022contrastive}. 
We argue that the existing InfoNCE relies on KL-divergence to measure the mutual information between variables of data augmentations.
Several deficiencies from KL-divergence limit the representation learning by InfoNCE.

\subsection{Limitations of KL Divergence}
As mutual information estimation utilizes KL-divergence to measure the similarity of distribution, mutual information estimation shares the limitations of KL-divergence. 
We argue that there are limitations to KL-divergence, including 
the exponential need for sample size and training instability.


\subsubsection{Exponential Need of Sample Size}
As derived by \cite{mcallester2020formal, ozair2019wasserstein}, the mutual information estimation based on KL-divergence has the high-confidence lower bound on $N$ samples that cannot be larger than $O(\ln N)$. 
With the application in contrastive learning InfoNCE, we have a similar theorem for mutual information estimation.
\begin{theorem}
Let $p(x_a)$ and $p(x_b)$ be two user sequence augmented distributions, and $A$ denotes the set of augmented sequences with sample size $N$ from $p(x_a)$, and $B$ denotes the set of augmented sequences with sample size $N$ from $p(x_b)$, respectively. 
Let $\delta$ be the confidence bound and let $F(A, B, \delta)$ be a real-valued function with augmented sets $A$, $B$, and the confidence parameter $\delta$. 
With probability $1-\delta$, we have
\begin{equation}
    D_{\text{KL}}(p(x_a), p(x_b))\geq F(A, B, \delta),
\end{equation}
then 
with at least $1-4\delta$ probability that
\begin{equation}
    \ln N \geq F(A, B, \delta).
\end{equation}
\end{theorem}
As mutual information is measured by KL-divergence, from~\cite{mcallester2020formal, ozair2019wasserstein}, we can conclude that the mutual information bound is $N=\exp(I(x_a, x_b))$. 
The $N$ in contrastive learning denotes the batch size. 
Therefore, the high-confidence mutual information lower bound estimation requires exponential sample sizes, which also matches with the derivation from Eq.~(\ref{eq:infonce_mutual}).

\subsubsection{Training Instability}
As demonstrated in analysis in WGAN~\cite{arjovsky2017wasserstein}, KL-divergence and Jensen-Shannon divergence both encounter unstable vanishing gradients when distributions are non-overlapping. 
KL-divergence can be infinite when sampled data have small probabilities close to $0$. 
As defined in Eq.~(\ref{eq:infonce_mutual}), when $p(x^{u_i}_b)$ has sampled points with probabilities $p(x^{u_i}_b)\approx 0$, the infinite KL-divergence happens. 
This case happens when the randomness of augmentations is large or user sequences are easily broken. 
Thus, it might cause training instability in mutual information estimation. 

\subsection{Wasserstein Discrepancy Measurement}
With these three limitations of KL-divergence, it is desirable to propose an alternative to KL-divergence in mutual information estimation. 
In this research, we propose Wasserstein Discrepancy Measurement in mutual information estimation. 
Formally, we define the Wasserstein Discrepancy Measurement with the negative 2-Wasserstein distance as follows:
\begin{align}
\label{eq:mutual_discrepancy_measurement}
    I_{W_2}\left(x^{u_i}_a, x^{u_i}_b\right) \stackrel{\text{def}}{=} -W_2(x^{u_i}_a, x^{u_i}_b) \propto \frac{p(x^{u_i}_a|x^{u_i}_b)}{p(x^{u_i}_b)},
\end{align}
where $-W_2(x^{u_i}_a, x^{u_i}_b)$ measures the negative 2-Wasserstein distribution distance between $\mathcal{N}(\mathbf{E}^{\mu}_{x^{u_i}_a}, \text{diag}(\mathbf{E}^{\Sigma}_{x^{u_i}_a}))$ and $\mathcal{N}(\mathbf{E}^{\mu}_{x^{u_i}_b}, \text{diag}(\mathbf{E}^{\Sigma}_{x^{u_i}_b}))$. 
2-Wasserstein distribution distance measures information gain from the metric learning perspective.
Wasserstein Discrepancy Measurement measures the negative optimal transport cost~\cite{chen2020graph} between augmentation distributions, which helps stabilize the gradient calculation and alleviates the training instability limitation. 
Moreover, 2-Wasserstein distance is symmetric, which indicates $W_2(x^{u_i}_a, x^{u_i}_b)=W_2(x^{u_i}_b, x^{u_i}_a)$, and further demonstrates less need of batch size in estimating mutual information, compared with KL-divergence. 

\section{Mutual Wasserstein Discrepancy Minimization}
Considering the stochasticity of data augmentations with stochastic modeling, we propose Wasserstein Discrepancy Measurement in the InfoNCE framework. 
We minimize Wasserstein discrepancy measurement $\mathcal{L}_{\text{MStein}}$ (equivalent to maximizing the mutual information $I_{W_2}\left(x^{u_i}_a, x^{u_i}_b\right)$) as follows:
\begin{align}
\label{eq:mstein}
    &\mathbb{E}_{\mathcal{S}^{u_i}\in\mathcal{B}}I_{W_2}\left(x^{u_i}_a, x^{u_i}_b\right) \geq \mathbb{E}_{\mathcal{S}^{u_i}\in\mathcal{B}}-\mathcal{L}_{\text{MStein}}\left(\textbf{h}^{u_i}_a, \textbf{h}^{u_i}_b\right)\nonumber\\
    &= \mathbb{E}_{\mathcal{S}^{u_i}\in\mathcal{B}}\log\frac{\exp\left(-W_2(\textbf{h}^{u_i}_a, \textbf{h}^{u_i}_b)\right)}{\exp\left(-W_2(\textbf{h}^{u_i}_a, \textbf{h}^{u_i}_b)\right)+\sum_{j\in S_{\mathcal{B}}^-}\exp\left(-W_2(\textbf{h}^{u_i}_a, \textbf{h}^j)\right)},
\end{align}
where $\left(\textbf{h}^{u_i}_{\mu},\textbf{h}^{u_i}_{\Sigma}\right) = \text{StosaEnc}(\mathcal{S}^{u_i})$ are encoded stochastic output representations, and the 2-Wasserstein distance on encoded distribution is $-W_2(\textbf{h}^{u_i}_a, \textbf{h}^{u_i}_b)=-\left(||\mu_{x^{u_i}_a}-\mu_{x^{u_i}_b}||^2_2 + ||\Sigma_{x^{u_i}_a}^{1/2}-\Sigma_{x^{u_i}_b}^{1/2}||^2_{\text{F}}\right)$, which is the sum of two $L2$-errors on both mean embeddings and the square root of covariance embeddings.
We measure the Wasserstein discrepancy of all augmented sequence pairs. 
The discrepancy is minimized for positive pairs, while the discrepancy is maximized for negative pairs. 
With $\mathcal{L}_{\text{MStein}}$, both stochasticities of augmentations and sequential behaviors are modeled. 
Moreover, adopting the 2-Wasserstein distance to measure the mutual information requires less batch size with symmetric estimation and more stable training. 

\subsection{Approixmating Lipschitz Continuity for Robustness}
The stable training stability originates from the approximating Lipschitz continuity of STOSA and \modelname. 
Intuitively, a model is Lipschitz continuous when a certain amount of inputs bounds its embedding output with no more than Lipschitz constant times that amount~\cite{kim2021lipschitz}. 
Lipschitz continuity is closely related to the robustness of the model against perturbations, which is also a necessary component in contrastive learning robustness. 
We utilize the demonstration from \cite{kim2021lipschitz} that the dot-product self-attention module is not Lipschitz, but the self-attention based on the $L2$ norm is Lipschitz instead. 
The 2-Wasserstein distance with the diagonal covariance is the sum of two $L2$ errors on both mean embeddings and the square root of covariance embeddings. 
The Wasserstein self-attention proposed by STOSA approximates Lipschitz. 
Moreover, the approximated Lipschitz continuity of the encoder further derives the Lipschitz approximation of the proposed $\mathcal{L}_{\text{MStein}}$, which improves the robustness. 
We empirically demonstrate the robustness of \modelname in experiment Section \ref{sec:robustness}. In the actual implementation, we relax the requirement that $W_Q=W_K$ in the Wasserstein self-attention module to approximate Lipschitz continuity for better flexibility and better performances.

\subsection{Exact Optimization of Alignment and Uniformity}
We further show that mutual Wasserstein discrepancy minimization exactly optimizes two important properties, alignment and uniformity~\cite{wang2020understanding}. Specifically, by decomposing the $\mathcal{L}_{\text{MStein}}(\textbf{h}^{u_i}_a, \textbf{h}^{u_i}_b)$, we obtain the alignment component from the nominator of Eq.~(\ref{eq:mstein}) as:
\begin{align}
\label{eq:alignment}
    ||\mu_{x^{u_i}_a}-\mu_{x^{u_i}_b}||^2_2 + ||\Sigma_{x^{u_i}_a}^{1/2}-\Sigma_{x^{u_i}_b}^{1/2}||^2_{\text{F}},
\end{align}
and the uniformity component from the denominator of Eq.~(\ref{eq:mstein}) as:
\begin{align}
\label{eq:uniformity}
    &-\log \sum\exp\left(-W_2(\textbf{h}^{u_i}_a, \textbf{h}^{u_j}_b)\right)\nonumber\\
    &=\log \sum \exp\left(||\mu_{x^{u_i}_a}-\mu_{x^{u_j}_b}||^2_2\right) \cdot \exp\left(||\Sigma_{x^{u_i}_a}^{1/2}-\Sigma_{x^{u_j}_b}^{1/2}||^2_{\text{F}}\right)\nonumber\\
    &=\log \sum \exp\left(||\mu_{x^{u_i}_a}-\mu_{x^{u_j}_b}||^2_2\right) + \log \sum \exp\left(||\Sigma_{x^{u_i}_a}^{1/2}-\Sigma_{x^{u_j}_b}^{1/2}||^2_{\text{F}}\right)
\end{align}

{\bf Commonality: } Both Eq.~(\ref{eq:alignment}) and Eq.~(\ref{eq:uniformity}) have similar forms as the original alignment and uniformity as defined in~\cite{wang2020understanding,qiu2022contrastive}. The Eq.~(\ref{eq:alignment}) also adopts the Euclidean distance between embeddings (alignment on representations) and the Eq.~(\ref{eq:uniformity}) also adopts the exponential Euclidean distance on all pairs of augmented sequences (uniformity on representations).

{\bf Differences and Novelty: } The differences between our proposed $\mathcal{L}_{\text{MStein}}(\textbf{h}^{u_i}_a, \textbf{h}^{u_i}_b)$ and \cite{wang2020understanding,qiu2022contrastive} from two perspectives: (1). We introduce the alignment and uniformity optimizations also on the covariance embeddings, with the advantage of pulling similar users' augmentation distributions together (\textit{i.e.,} \textbf{distribution alignment}) and enforcing the distributions to be as distinguishable as possible (\textit{i.e.,} \textbf{distribution uniformity}); (2). the alignment and uniformity terms proposed in~\cite{wang2020understanding} are \textbf{asymptotically} optimized by the contrastive loss and are not induced from the original formulation of the contrastive loss. However, our proposed $\mathcal{L}_{\text{MStein}}(\textbf{h}^{u_i}_a, \textbf{h}^{u_i}_b)$ induces and optimizes \textbf{exactly} the alignment and uniformity terms. In other words, the alignment and uniformity optimizations proposed in~\cite{wang2020understanding} and the Euclidean metric used in CL by~\cite{ozair2019wasserstein} can be viewed as a special case of our $\mathcal{L}_{\text{MStein}}(\textbf{h}^{u_i}_a, \textbf{h}^{u_i}_b)$, which adopts Euclidean distance instead of Wasserstein distance. 

\subsection{Optimization and Prediction}
With mutual Wasserstein discrepancy minimization $\mathcal{L}_{\text{MStein}}$, we finalize the optimization loss with the recommendation loss from Eq.~(\ref{eq:rec_loss}) as follows:
\begin{align}
    \mathcal{L} = \mathcal{L}_{rec} + \beta\mathcal{L}_{\text{MStein}},
\end{align}
where $\beta$ is the hyper-parameter for adjusting the contribution of contrastive loss with mutual Wasserstein discrepancy minimization. The final recommendation list is generated by calculating the Wasserstein distance of the sequence encoded distribution embeddings $(\textbf{h}_u^{\mu},\textbf{h}_u^{\Sigma})$ and all items' stochastic embeddings. The distances on all items are sorted in the ascending order to produce the top-N.

\section{Experiments}
In this section, we demonstrate the effectiveness of the proposed \modelname in multiple aspects, including performances over baselines, robustness against perturbations, and analysis of performances over different batch sizes. We answer the following research questions~(RQs) in experiments:
\begin{itemize}[leftmargin=*]
    \item \textbf{RQ1: }Is \modelname generating better recommendations than state-of-the-art baselines?
    \item \textbf{RQ2: }Is \modelname more robust to noisy and limited data?
    \item \textbf{RQ3: }Does \modelname need smaller batch sizes?
    \item \textbf{RQ4: }Where are improvements of \modelname from?
\end{itemize}

\subsection{Baselines}
We compare the proposed \modelname with three groups of recommendation methods. The first group includes static recommendation methods. We present BPRMF~\cite{rendle2012bpr} due to the page limitation. The second group of methods include state-of-the-art sequential recommendation methods without self-supervised module, including Caser~\cite{tang2018personalized}, SASRec~\cite{kang2018self}, BERT4Rec~\cite{sun2019bert4rec}, and STOSA~\cite{fan2022sequential}. The third group contains most recent sequential recommendation methods with self-supervised learning, including CL4Rec~\cite{xie2022contrastive}, DuoRec~\cite{qiu2022contrastive}, and CoSeRec~\cite{liu2021contrastive}. We also introduce a variant that builds upon CL methods with SASRec as base backbone but uses WDM as CL loss, which is CoSeRec(WDM) by converting the sequence output embeddings as $[mean\_emb;ELU(cov\_emb)+1]$. Note that we use only \textbf{one training negative sample} for models with the Cross-Entropy loss~(\textit{e.g.,} DuoRec) because we observe that the number of negative samples significantly affects the recommendation performance~\cite{ding2019reinforced, chen2017sampling, mao2021simplex}. 

\begin{table*}[]
\centering
\caption{Overall Performance Comparison Table. The best results are bold and the best baseline results are underlined, respectively. `Improve.' indicates the relative improvement against the best baseline performance.}
\label{tab:over_perf}
\resizebox{\textwidth}{!}{%
\begin{tabular}{@{}ccccccccccccc@{}}
\toprule
Dataset & Metric & BPRMF & Caser & SASRec & BERT4Rec & STOSA & CL4Rec & DuoRec & \multicolumn{1}{l}{CoSeRec} & CoSeRec(WDM) & \modelname & Improv. \\ \midrule
\multirow{6}{*}{Beauty} & Recall@1 & 0.0082 & 0.0112 & 0.0129 & 0.0119 & {\ul 0.0193} & 0.0156 & 0.0158 & 0.0188  & 0.0189 & \textbf{0.0220} & \multicolumn{1}{l}{+14.39\%} \\
 & Recall@5 & 0.0300 & 0.0309 & 0.0416 & 0.0396 & 0.0504 & {\ul 0.0538} & 0.0505 & 0.0508 & 0.0524 & \textbf{0.0551} & +2.24\% \\
 & NDCG@5 & 0.0189 & 0.0214 & 0.0274 & 0.0257 & {\ul 0.0351} & 0.0349 & 0.0310 & {\ul 0.0351} & 0.0359 & \textbf{0.0392} & +11.69\% \\
 & Recall@10 & 0.0471 & 0.0407 & 0.0633 & 0.0595 & 0.0707 & 0.0726 & 0.0685 & {\ul 0.0738} & 0.0760 & \textbf{0.0774} & +4.78\% \\
 & NDCG@10 & 0.0245 & 0.0246 & 0.0343 & 0.0321 & 0.0416 & 0.0412 & 0.0375 & {\ul 0.0425} & 0.0435 & \textbf{0.0463} & +9.00\% \\
 & MRR & 0.0216 & 0.0231 & 0.0291 & 0.0294 & 0.0360 & 0.0356 & 0.0325 & {\ul 0.0365} & 0.0368 & \textbf{0.0398} & +9.11\% \\ \midrule
\multirow{6}{*}{Tools} & Recall@1 & 0.0062 & 0.0056 & 0.0103 & 0.0059 & {\ul 0.0120} & 0.0112 & 0.0108 & 0.0112 & 0.0114 & \textbf{0.0144} & \multicolumn{1}{l}{+20.10\%} \\
 & Recall@5 & 0.0216 & 0.0129 & 0.0284 & 0.0189 & 0.0312 & 0.0314 & 0.0304 & {\ul 0.0318} & \textbf{0.0344} & 0.0334 & +8.17\% \\
 & NDCG@5 & 0.0139 & 0.0091 & 0.0194 & 0.0123 & {\ul 0.0217} & 0.0208 & 0.0201 & 0.0216 & 0.0230 & \textbf{0.0242} & +11.11\% \\
 & Recall@10 & 0.0334 & 0.0193 & 0.0427 & 0.0319 & {\ul 0.0468} & 0.0404 & 0.0401 & 0.0453 & \textbf{0.0487} & 0.0472 & +4.06\% \\
 & NDCG@10 & 0.0177 & 0.0112 & 0.0240 & 0.0165 & {\ul 0.0267} & 0.0226 & 0.0234 & 0.0260 & 0.0276 & \textbf{0.0286} & +6.90\% \\
 & MRR & 0.0154 & 0.0106 & 0.0207 & 0.0160 & {\ul 0.0226} & 0.0212 & 0.0202 & 0.0223 & 0.0234 & \textbf{0.0248} & +9.90\% \\ \midrule
\multirow{6}{*}{Toys} & Recall@1 & 0.0084 & 0.0089 & 0.0193 & 0.0110 & {\ul 0.0240} & 0.0220 & 0.0215 & 0.0222 & 0.0228 & \textbf{0.0266} & \multicolumn{1}{l}{+10.73\%} \\
 & Recall@5 & 0.0301 & 0.0240 & 0.0551 & 0.0300 & 0.0577 & {\ul 0.0617} & 0.0580 & 0.0584 & 0.0616 & \textbf{0.0637} & +3.17\% \\
 & NDCG@5 & 0.0194 & 0.0210 & 0.0377 & 0.0206 & 0.0412 & {\ul 0.0424} & 0.0401 & 0.0408 & 0.0426 & \textbf{0.0457} & +7.78\% \\
 & Recall@10 & 0.0460 & 0.0262 & 0.0797 & 0.0466 & {\ul 0.0800} & 0.0764 & 0.0784 & 0.0791 & \textbf{0.0852} & 0.0845 & +6.50\% \\
 & NDCG@10 & 0.0245 & 0.0231 & 0.0456 & 0.0260 & {\ul 0.0481} & 0.0454 & 0.0461 & 0.0474 & 0.0502 & \textbf{0.0524} & +8.91\% \\
 & MRR & 0.0216 & 0.0221 & 0.0385 & 0.0244 & 0.0415 & {\ul 0.0417} & 0.0400 & 0.0405 & 0.0425 & \textbf{0.0453} & +8.67\% \\ \midrule
\multirow{6}{*}{Office} & Recall@1 & 0.0073 & 0.0069 & 0.0198 & 0.0137 & 0.0234 & 0.0230 & 0.0221 & {\ul 0.0245} & 0.0267 & \textbf{0.0277} & \multicolumn{1}{l}{+13.33\%} \\
 & Recall@5 & 0.0214 & 0.0302 & 0.0656 & 0.0485 & 0.0677 & 0.0709 & 0.0665 & {\ul 0.0718} & 0.0703 & \textbf{0.0740} & +3.13\% \\
 & NDCG@5 & 0.0144 & 0.0186 & 0.0428 & 0.0309 & 0.0461 & 0.0471 & 0.0456 & {\ul 0.0483} & 0.0485 & \textbf{0.0512} & +5.93\% \\
 & Recall@10 & 0.0306 & 0.0550 & 0.0989 & 0.0848 & 0.1021 & 0.1091 & 0.1005 & {\ul 0.1024} & 0.1052 & \textbf{0.1155} & +5.96\% \\
 & NDCG@10 & 0.0173 & 0.0266 & 0.0534 & 0.0426 & 0.0572 & 0.0594 & 0.0556 & {\ul 0.0598} & 0.0597 & \textbf{0.0627} & +4.90\% \\
 & MRR & 0.0162 & 0.0268 & 0.0457 & 0.0408 & 0.0502 & 0.0511 & 0.0482 & {\ul 0.0516} & 0.0519 & \textbf{0.0529} & +2.53\% \\ \bottomrule
\end{tabular}
}
\end{table*}
\subsection{Overall Comparisons~(RQ1)}
As demonstrated in the overall comparison results Table~\ref{tab:over_perf}, we can conclude the superiority of \modelname over all baselines in all metrics. We have the following observations:
\begin{itemize}[leftmargin=*]
    \item Among all models, the proposed \modelname achieves the consistently best performance in all metrics over all evaluated datasets. The improvements range from 0.9\% to 20.10\% in all metrics, proving the effectiveness of \modelname in SR. In the most challenging task top-1 recommendation, \modelname obtains the most significant improvements. In the entire list ranking metric MRR, \modelname achieves 2.53\% to 9.90\% improvements over the best baseline. We attribute these improvements to several characteristics of \modelname: (1). a novel mutual information estimation based on the 2-Wasserstein distance; (2). the uncertainty modeling for stochastic data augmentation processes in self-supervised learning; (3). the robust modeling from WDM. 
    \item Comparing the self-supervised learning SR methods~(CL4Rec, DuoRec, and CoSeRec), \modelname still achieves significant improvements among them. Although \modelname adopts the same data augmentations as CoSeRec, the performance improvements stem from the stochastic modeling of data augmentations and more accurate and robust mutual information estimation. Furthermore, CL4Rec and CoSeRec generate better performances among these baselines as both provide manually designed data augmentations. These observations demonstrate the benefits of modeling the uncertainty of data augmentation processes and the proposed Wasserstein Discrepancy Measurement.  
    \item In static models and SR methods, SR methods outperform the static models. This observation demonstrates the necessity of sequential information in recommendations. STOSA achieves the best performance in all SR methods, and the SASRec is the second best, showing that the self-attention module benefits SR. STOSA first introduces stochastic embeddings for modeling sequential uncertainty and demonstrates its effectiveness over other SR methods.
\end{itemize}

\subsection{Robustness Analysis~(RQ2)}
\label{sec:robustness}
We argue that \modelname is more robust with the newly proposed mutual Wasserstein discrepancy minimization process. We validate the robustness from two perspectives, including the robustness against noisy interactions and data sizes. The comparison is conducted in \modelname and CoSeRec because both adopt the same data augmentation techniques. 

\subsubsection{Sensitivity to Noisy Interactions}
\begin{figure}[]
\centering
     \begin{subfigure}[b]{0.235\textwidth}
         \centering
         \includegraphics[width=1\textwidth]{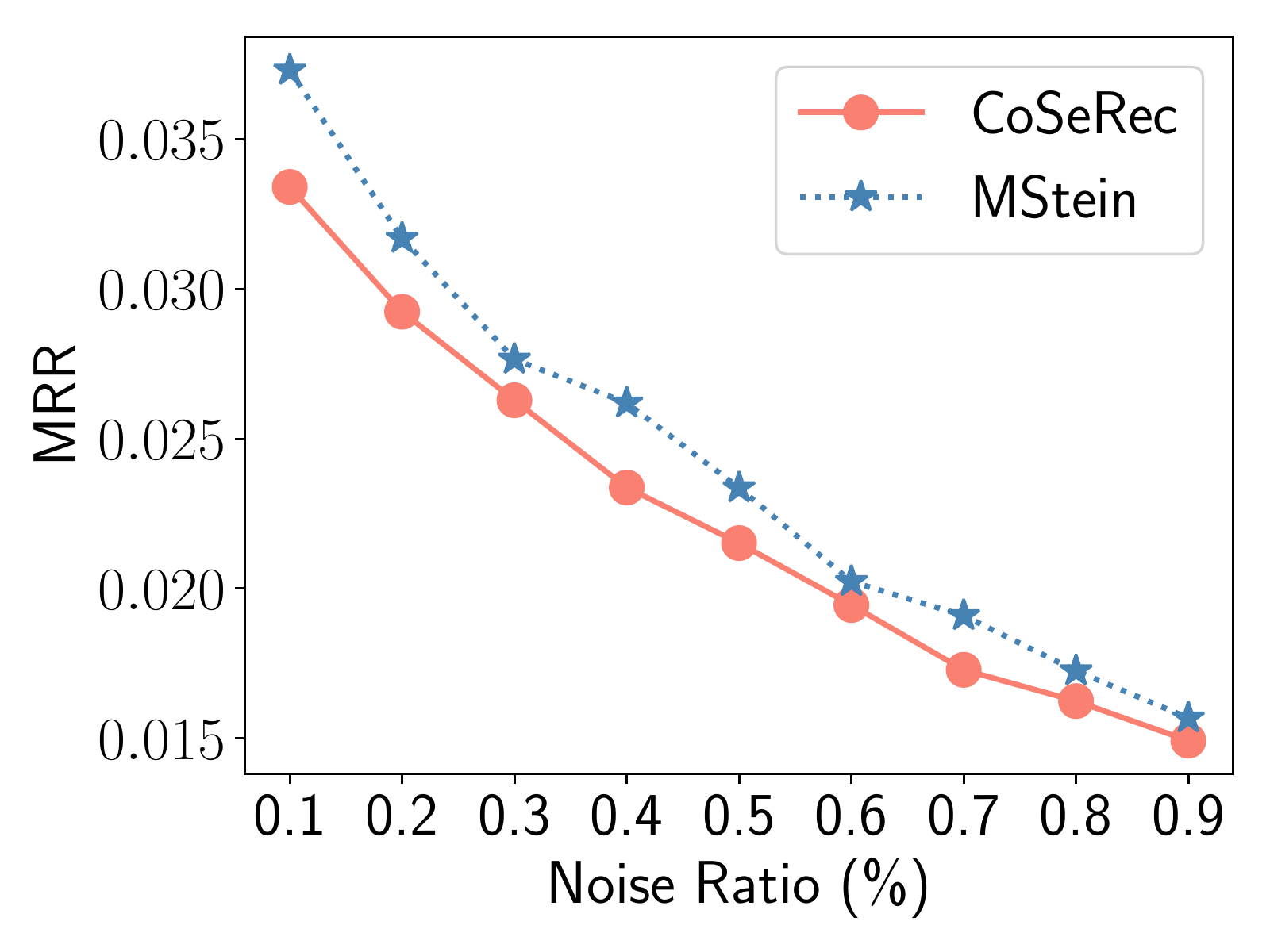}
         \caption{Beauty}
         \label{fig:beauty_noise}
     \end{subfigure}\hfill
     \begin{subfigure}[b]{0.235\textwidth}
         \centering
         \includegraphics[width=1\textwidth]{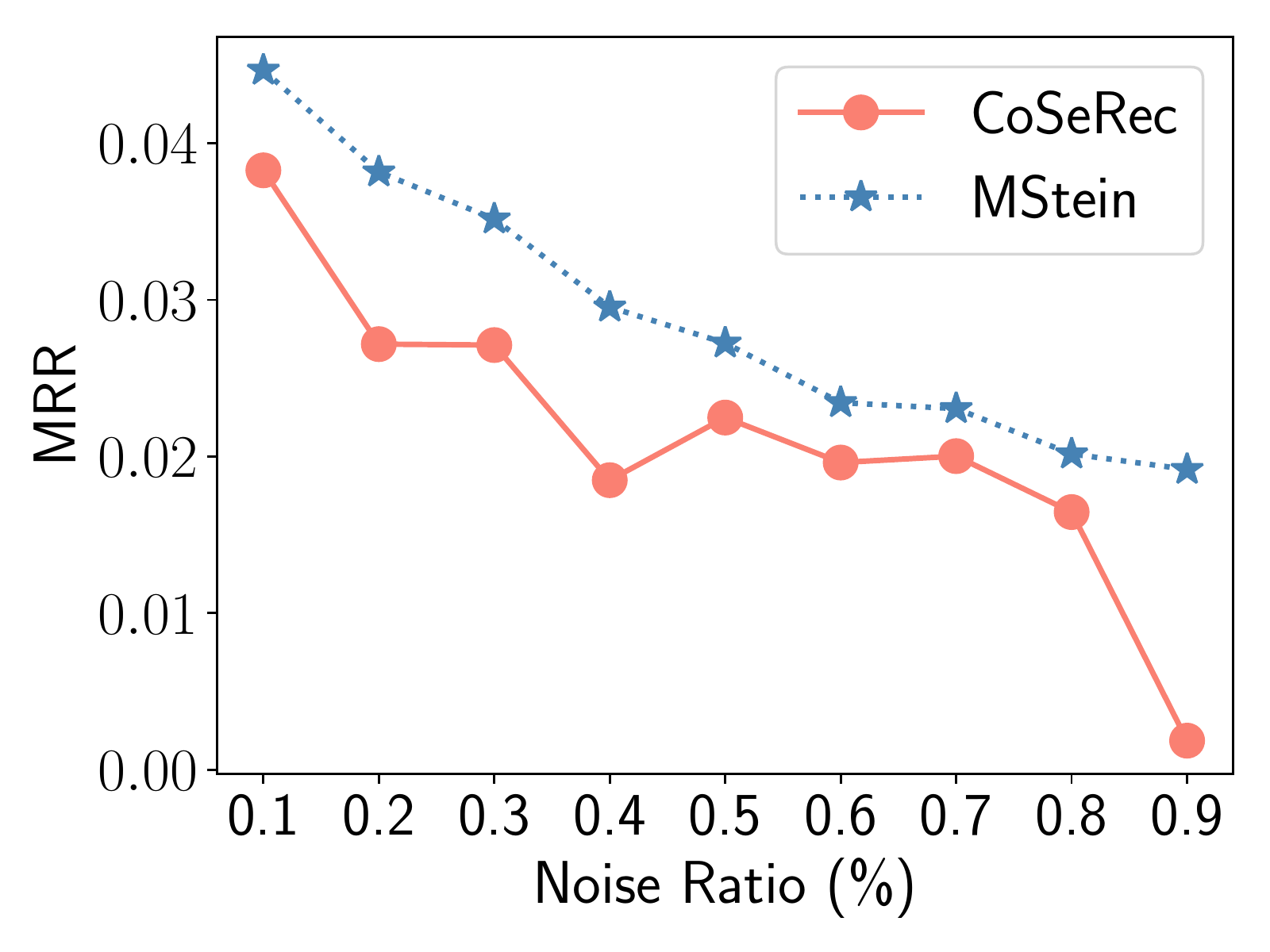}
         \caption{Toys}
         \label{fig:toys_noise}
     \end{subfigure}\\
     \begin{subfigure}[b]{0.235\textwidth}
         \centering
         \includegraphics[width=1\textwidth]{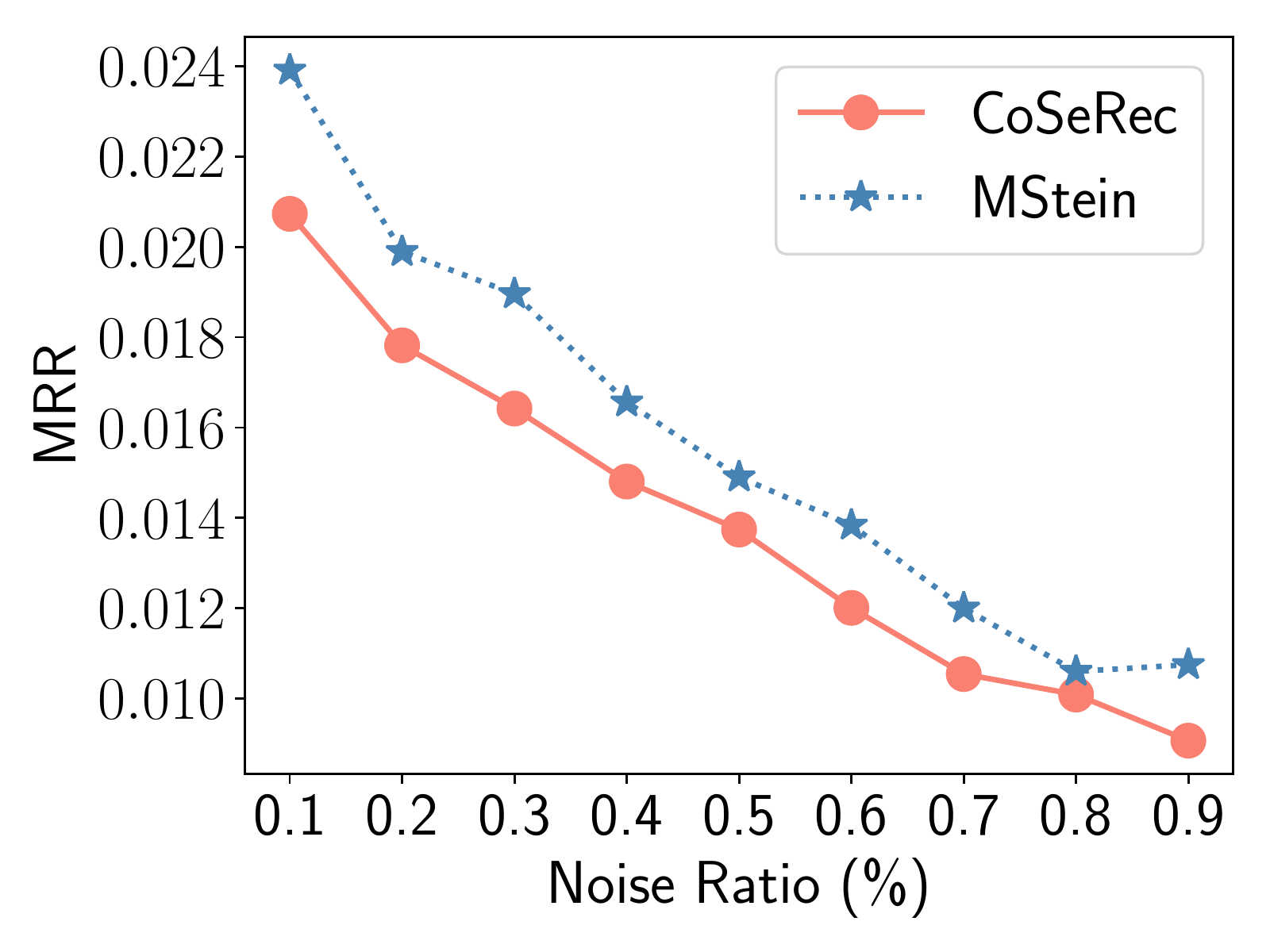}
         \caption{Tools}
         \label{fig:tools_noise}
     \end{subfigure}\hfill
     \begin{subfigure}[b]{0.235\textwidth}
         \centering
         \includegraphics[width=1\textwidth]{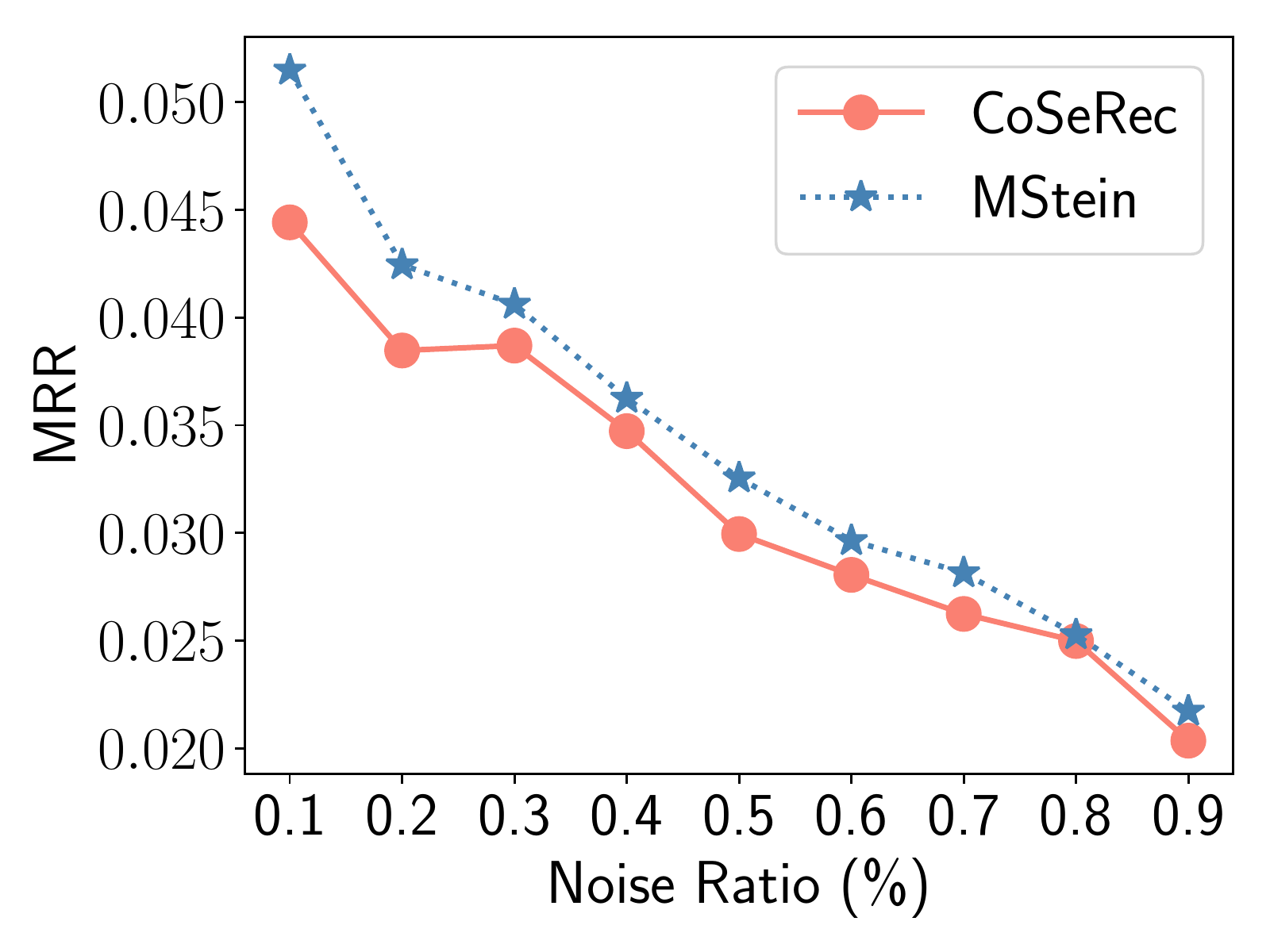}
         \caption{Office}
         \label{fig:office_noise}
     \end{subfigure}
\caption{MRR over Different Noise Ratios.}
\label{fig:mrr_over_noiseratio}
\end{figure}

We show the sensitivity analysis of \modelname against noisy interactions in Fig.~(\ref{fig:mrr_over_noiseratio}) in all datasets. Fig.~(\ref{fig:mrr_over_noiseratio}) shows the MRR performance over different noise ratios for the CoSeRec and the proposed \modelname. We can observe that \modelname is more robust to noisy interactions than CoSeRec. Specifically, for example, in the Beauty dataset analysis in Fig.~(\ref{fig:beauty_noise}), when the noise ratio is 0.4 for \modelname and 0.3 for CoSeRec,  the performances are similar. This observation shows the robustness of \modelname against noisy interaction with Wasserstein discrepancy measurement as \modelname and CoSeRec adopt the same data augmentation strategies. We can also see that the performance of CoSeRec drops significantly in the Toys dataset when the noise ratio is large~(0.9), while \modelname still achieves satisfactory performance.

\subsubsection{Sensitivity to Data Size}
\begin{figure}[]
\centering
     \begin{subfigure}[b]{0.235\textwidth}
         \centering
         \includegraphics[width=1\textwidth]{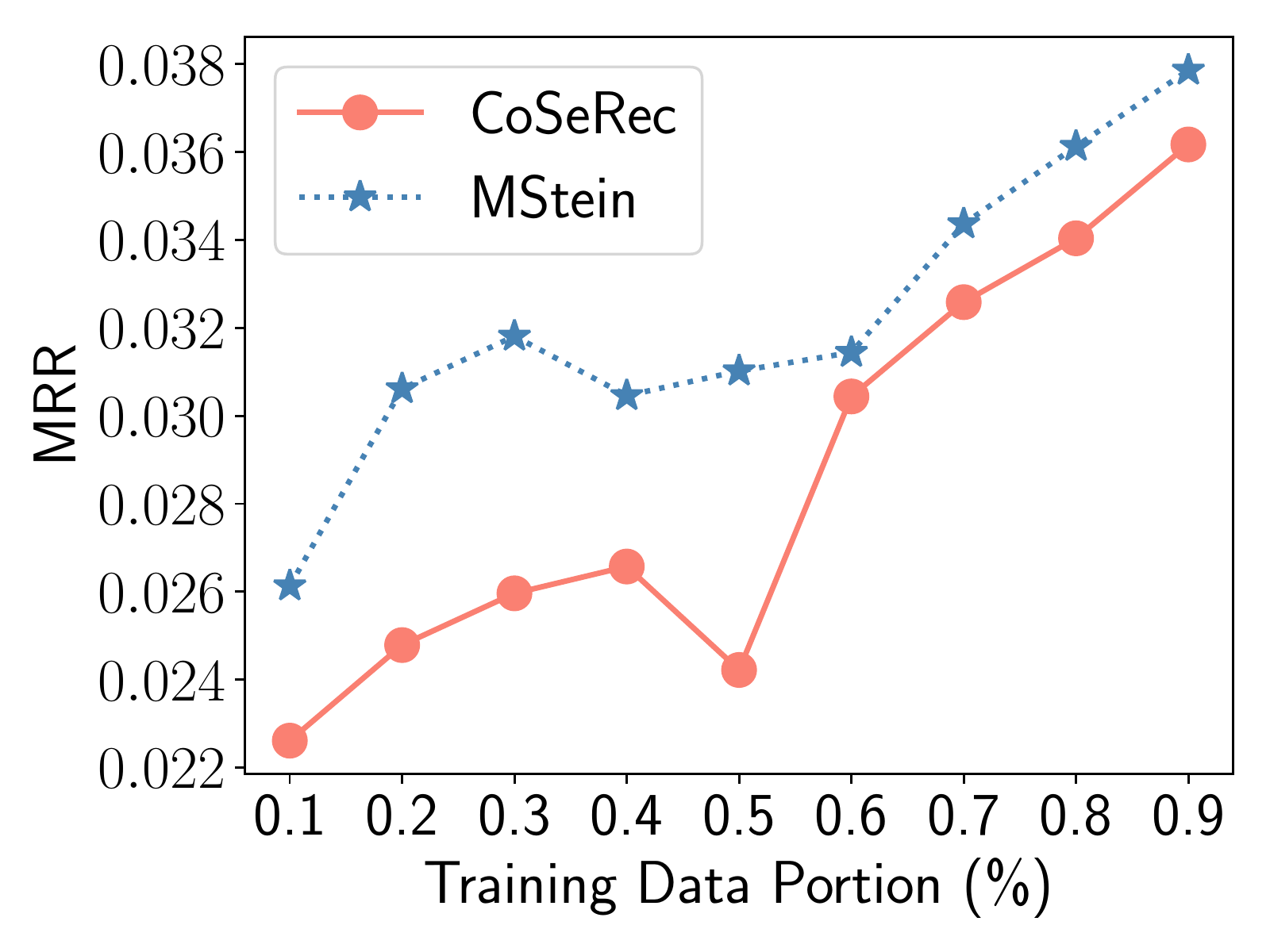}
         \caption{Beauty}
         \label{fig:beauty_trainportion}
     \end{subfigure}\hfill
     \begin{subfigure}[b]{0.235\textwidth}
         \centering
         \includegraphics[width=1\textwidth]{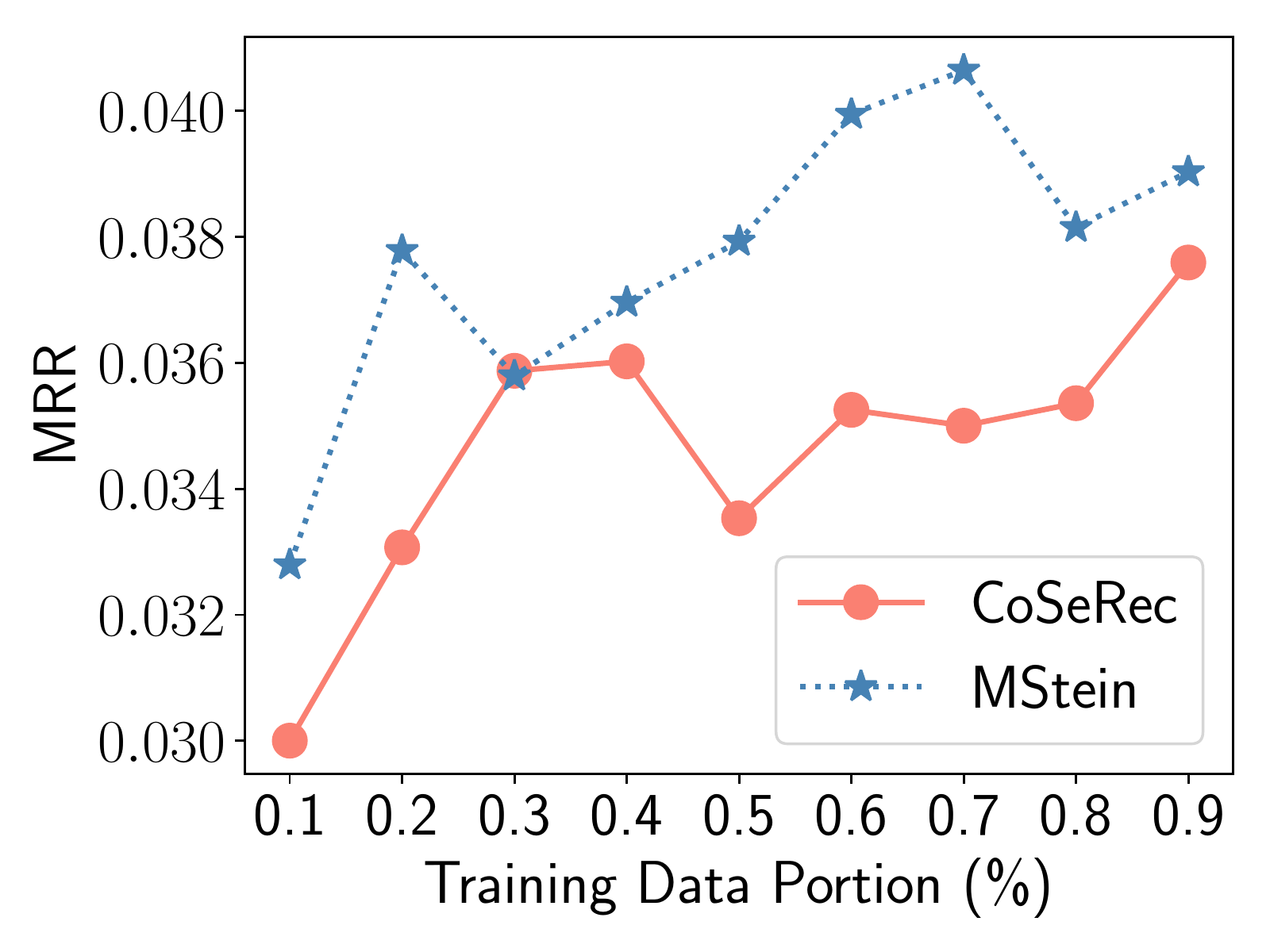}
         \caption{Toys}
         \label{fig:toys_trainportion}
     \end{subfigure}\\
     \begin{subfigure}[b]{0.235\textwidth}
         \centering
         \includegraphics[width=1\textwidth]{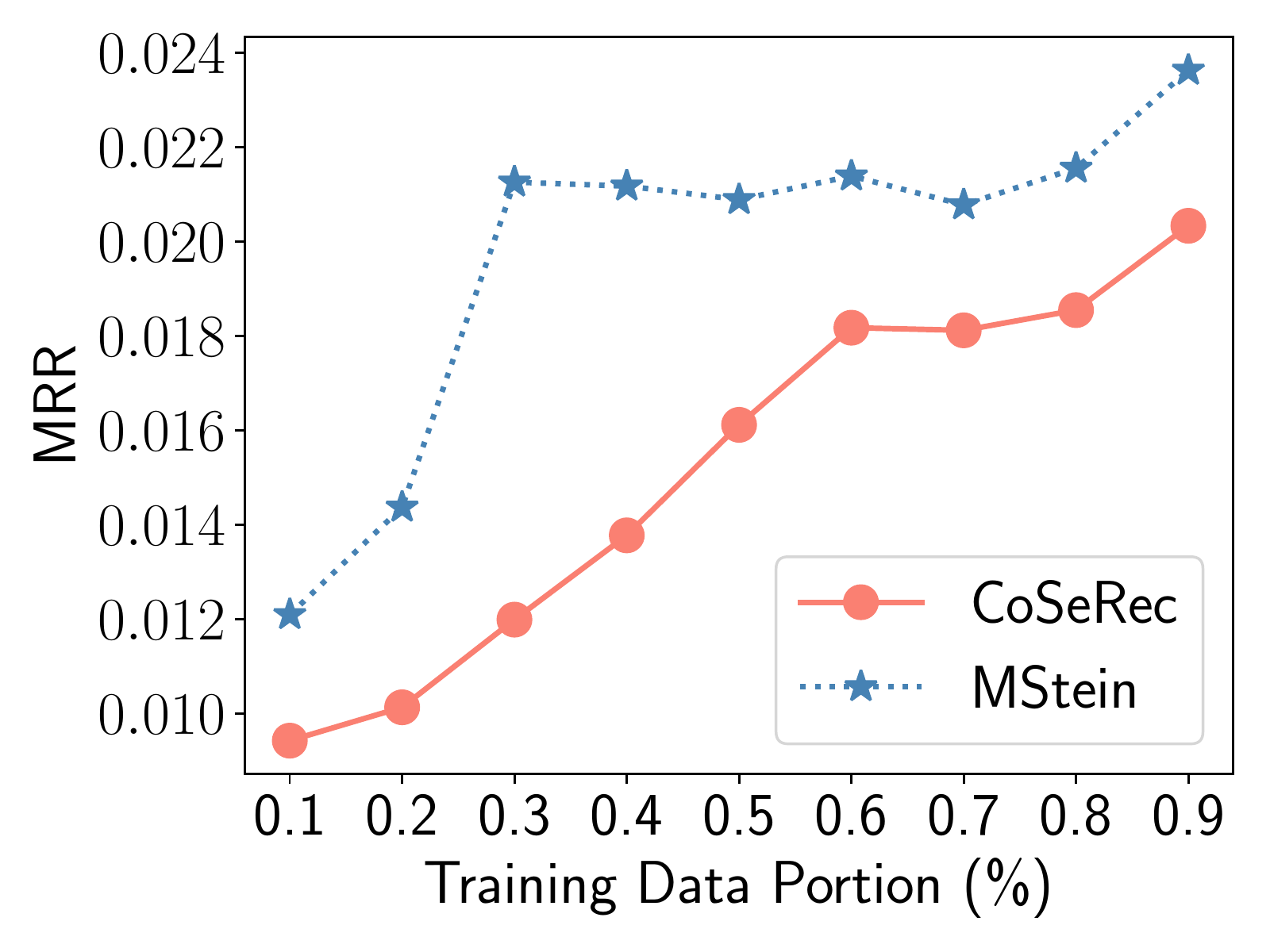}
         \caption{Tools}
         \label{fig:tools_trainportion}
     \end{subfigure}\hfill
     \begin{subfigure}[b]{0.235\textwidth}
         \centering
         \includegraphics[width=1\textwidth]{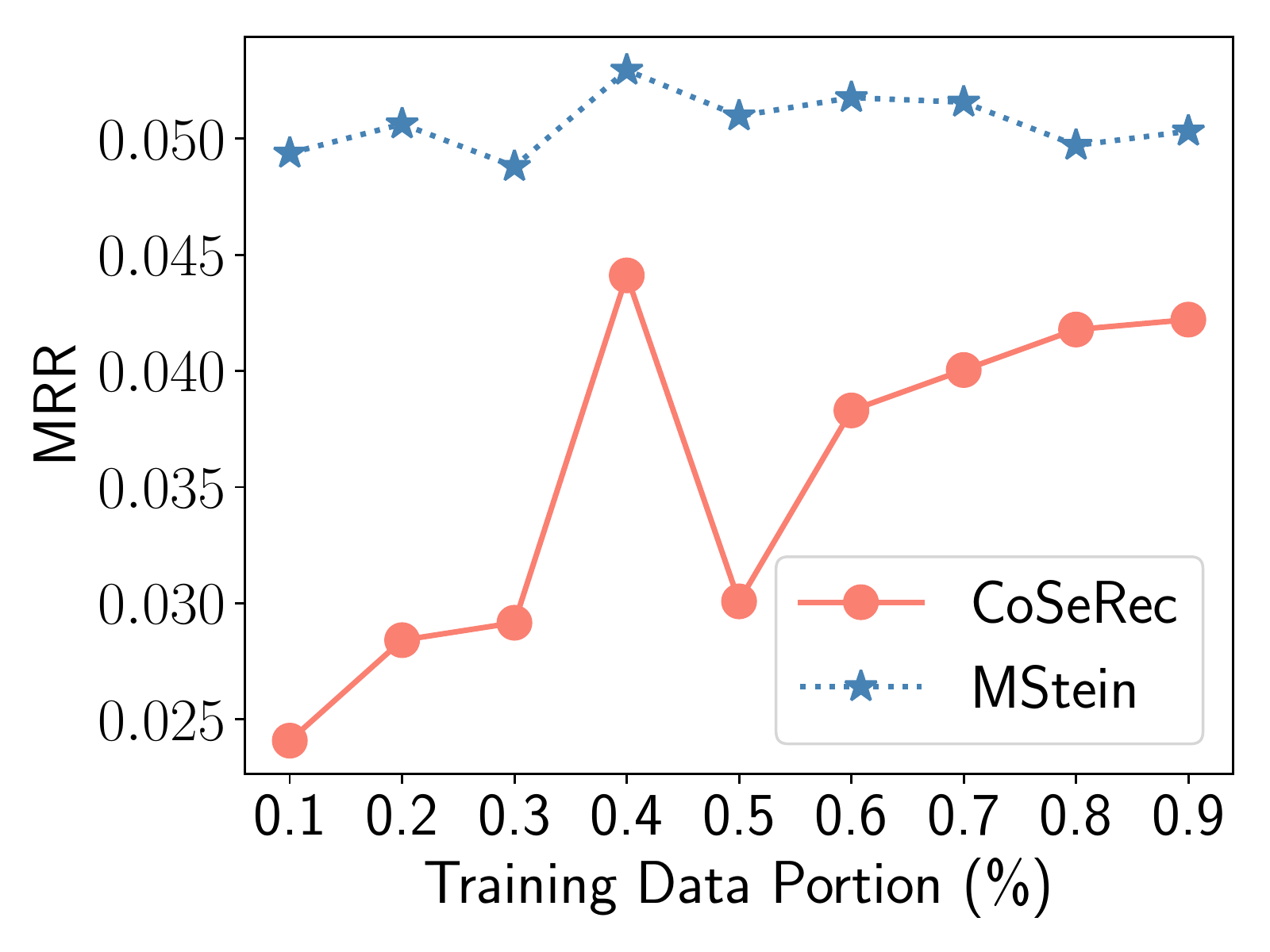}
         \caption{Office}
         \label{fig:office_trainportion}
     \end{subfigure}
\caption{MRR over Different Training Data Portions.}
\label{fig:mrr_over_trainportion}
\end{figure}

The sensitivity of \modelname against the data size is shown in Fig.~(\ref{fig:mrr_over_trainportion}). In Fig.~(\ref{fig:mrr_over_trainportion}), we present the performance comparison between \modelname and CoSeRec in varying data sizes. We can observe that \modelname consistently outperforms CoSeRec in all varying data size ratios, demonstrating the superiority of \modelname in SR. Moreover, \modelname is more stable than CoSeRec, especially in Tools and Office datasets, as shown in Fig.~(\ref{fig:tools_trainportion}) and Fig.~(\ref{fig:office_trainportion}) respectively. It demonstrates that \modelname is more robust than CoSeRec against the dataset size, potentially due to the collaborative transitivity from stochastic embedding modeling and the newly proposed Wasserstein discrepancy measurement.

\subsection{Sensitivity to Batch Size~(RQ3)}
\begin{figure}[]
\centering
     \begin{subfigure}[b]{0.235\textwidth}
         \centering
         \includegraphics[width=1\textwidth]{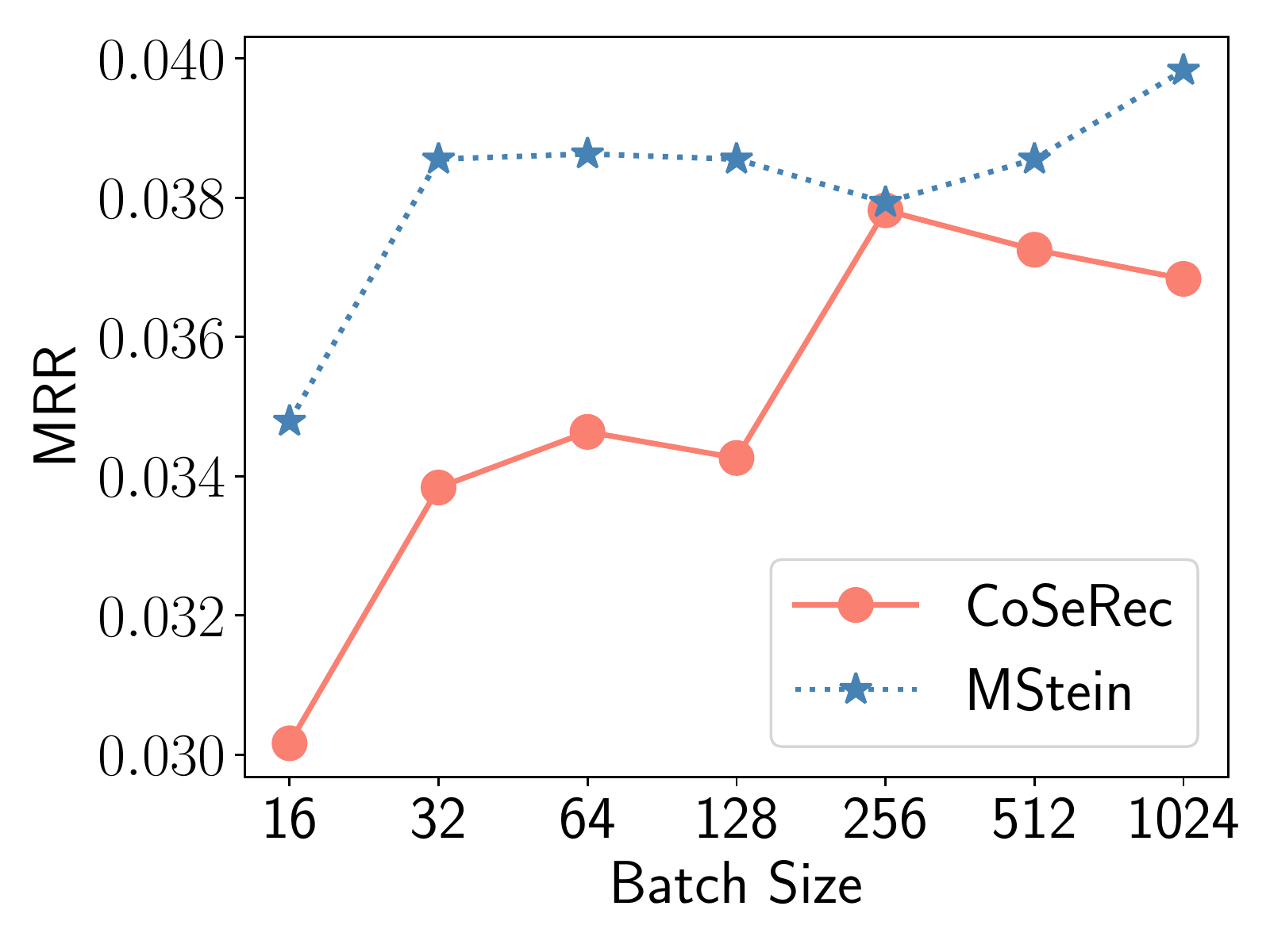}
         \caption{Beauty}
         \label{fig:beauty_batch}
     \end{subfigure}\hfill
     \begin{subfigure}[b]{0.235\textwidth}
         \centering
         \includegraphics[width=1\textwidth]{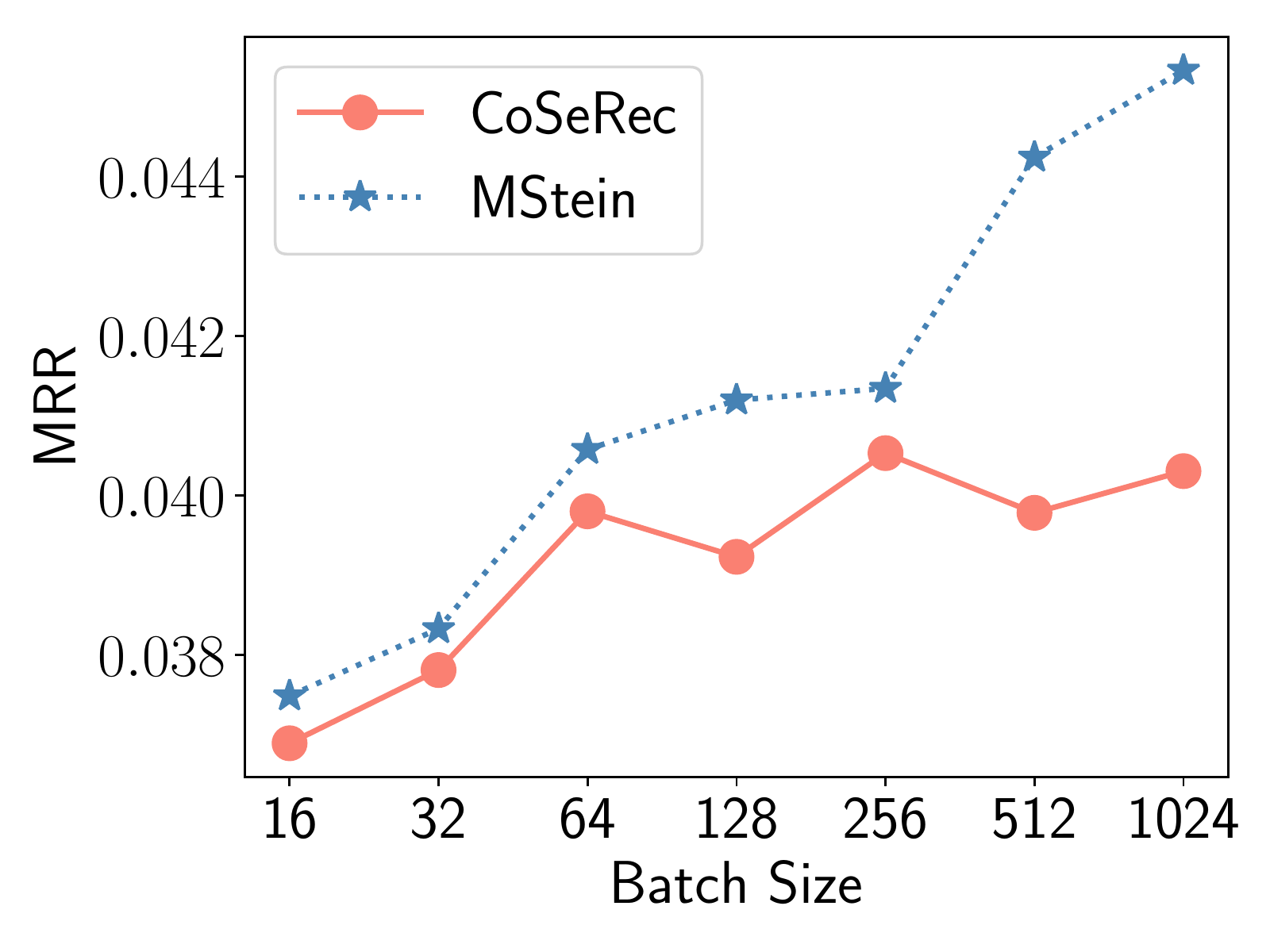}
         \caption{Toys}
         \label{fig:toys_batch}
     \end{subfigure}\\
     \begin{subfigure}[b]{0.235\textwidth}
         \centering
         \includegraphics[width=1\textwidth]{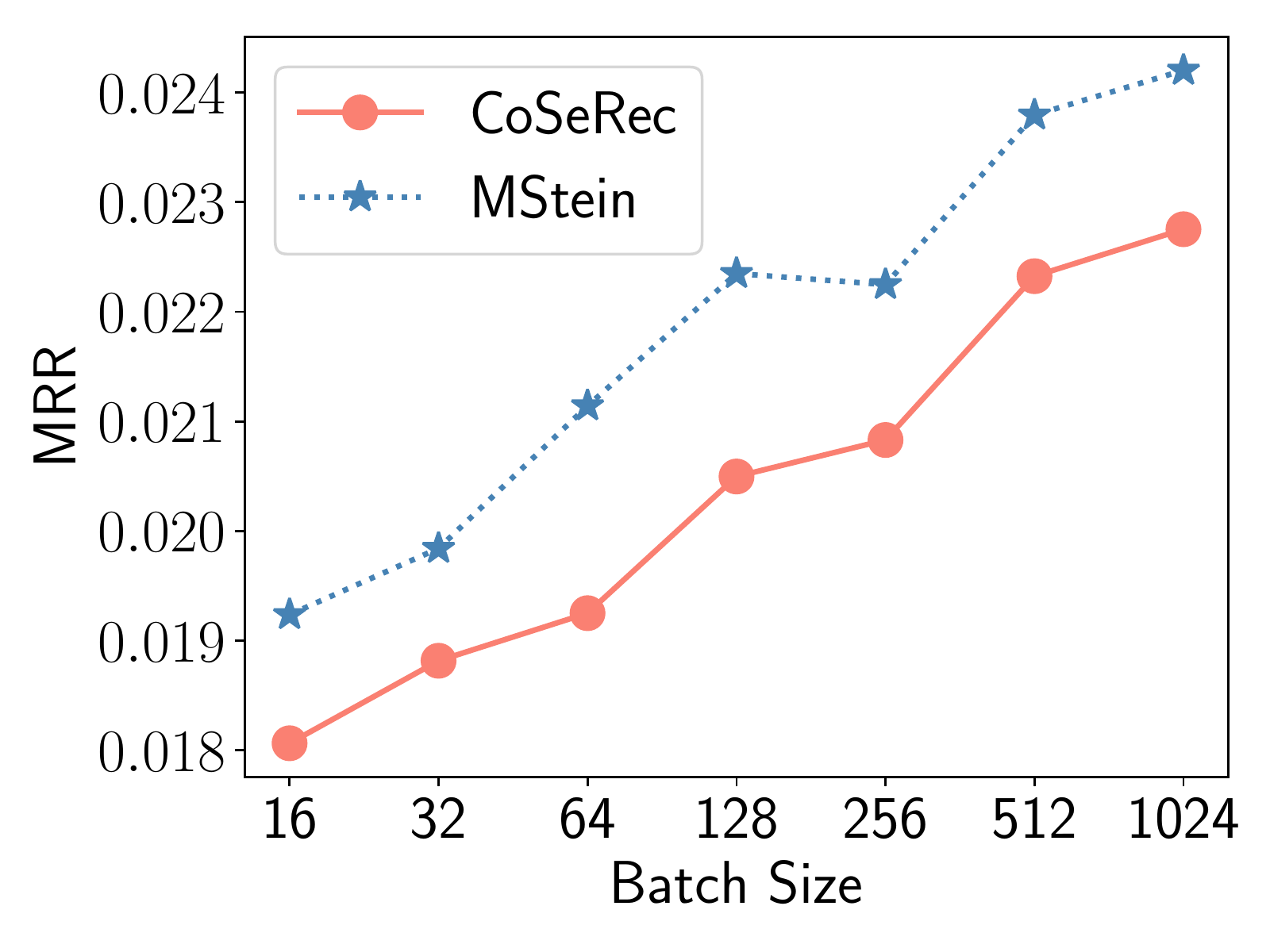}
         \caption{Tools}
         \label{fig:tools_batch}
     \end{subfigure}\hfill
     \begin{subfigure}[b]{0.235\textwidth}
         \centering
         \includegraphics[width=1\textwidth]{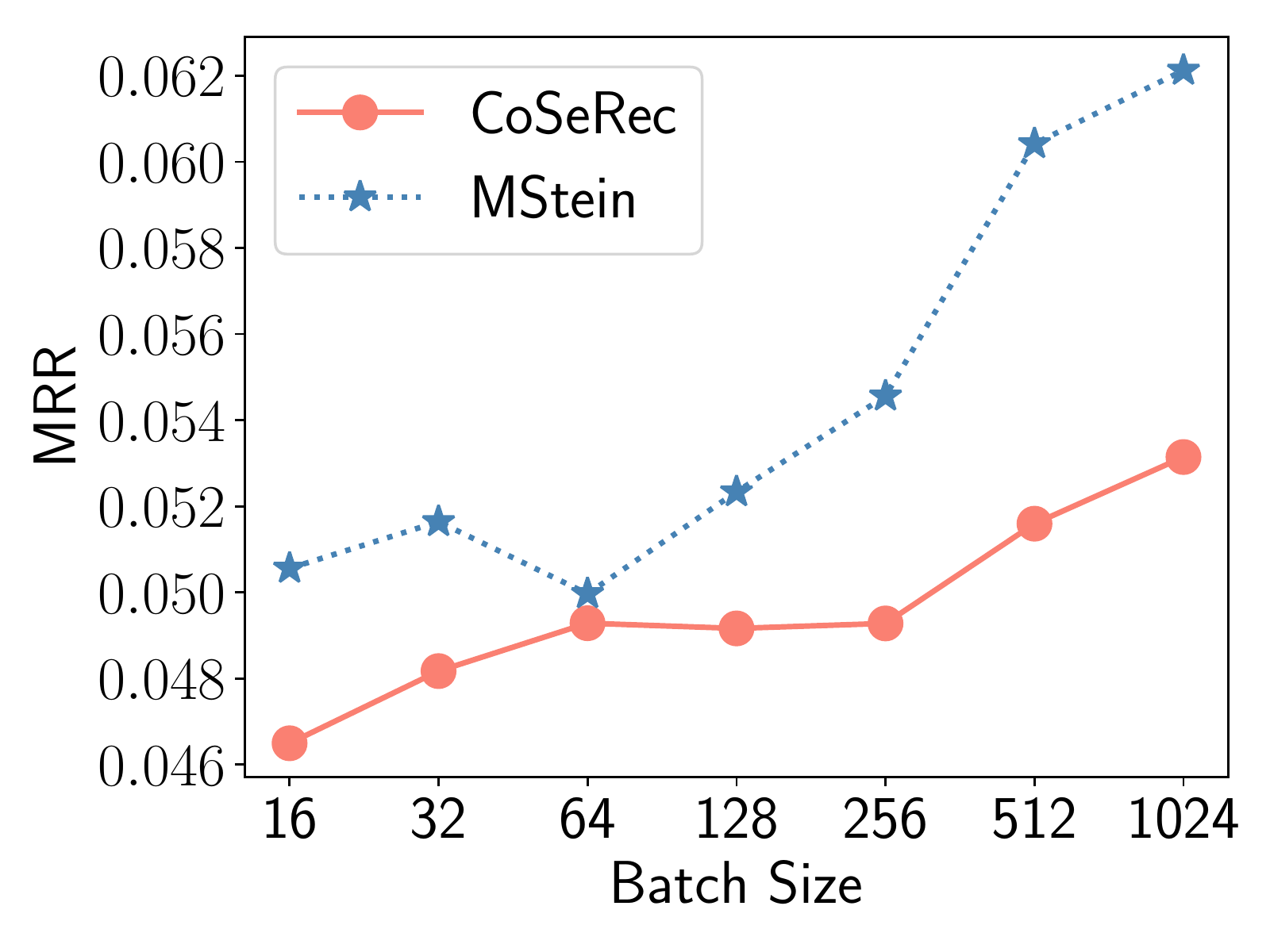}
         \caption{Office}
         \label{fig:office_batch}
     \end{subfigure}
\caption{MRR over Different Batch Sizes.}
\label{fig:mrr_overbatch}
\end{figure}

As we argue that the proposed Wasserstein discrepancy measurement alleviates the exponential need for the sample size of KL divergence in mutual information estimation, we conduct the sensitivity analysis to batch sizes in Fig.~(\ref{fig:mrr_overbatch}). We compare the proposed \modelname with CoSeRec in this analysis as the same set of data augmentations is applied. In contrastive learning, larger batch sizes improve the model performance significantly~\cite{oord2018representation}. Fig.~(\ref{fig:mrr_overbatch}) demonstrates the beneficial effect of using larger batch sizes. Moreover, in all four datasets, \modelname achieves comparative performances with much smaller batch sizes. For example, in Fig.~(\ref{fig:beauty_batch}), \modelname obtains MRR as 0.035 when batch size is 16~($2^4$) while CoSeRec needs the batch size 128~($2^7$). This observation validates the superiority of \modelname over CoSeRec in need of sample size, where CoSeRec needs exponential sample sizes in InfoNCE measurement due to the inherent KL divergence limitation. 

\subsection{Improvements Analysis~(RQ4)}\label{sec:imp_analys}
We visualize the improvements of users and items in Appendix Fig.~(\ref{fig:ndcg5_seqlen}) and items in Fig.~(\ref{fig:item_pop_full}) in on all datasets. We separate users and items in groups based on the number of interactions. For each group, we average NDCG@5 over the group users\slash items. We observe that the distributions of users\slash items based on the number of interactions follow the long-tail distributions shown in the bar chart. In most datasets, the performance increases as the number of interactions grow. The proposed \modelname achieves better performance than SASRec, STOSA, and CoSeRec. The improvements come from the groups with the longest sequences and the second longest sequences. It verifies the strength of \modelname for modeling stochastic augmentations because data augmentations for long sequences provide richer perspectives of sequences. For short sequences, data augmentations can easily break the sequential correlations. Long sequences and popular items have larger uncertainty, and stochastic augmentations provide more informative signals in contrastive learning. This observation also happens to the item perspective. \modelname also achieves better performance in popular items.

\section{Conclusions}
We study the connection between mutual information and InfoNCE and discuss the limitations of mutual information estimation based on KL-divergence, including asymmetrical estimation, the exponential need for sample size, and the training instability. We propose an alternative choice of mutual information estimation based on Wasserstein distance, which is Wasserstein Discrepancy Measurement. With the proposed Wasserstein Discrepancy Measurement, we formulate the mutual Wasserstein discrepancy minimization in the InfoNCE framework as \modelname. Extensive experiments on four benchmark datasets demonstrate the superiority of \modelname using Wasserstein Discrepancy Measurement in mutual information estimation. Additional robustness analysis proves that \modelname is more robust against noisy interactions and variants of data sizes. 

\begin{acks}
This paper was supported by the National Key R\&D Program of China through grant 2022YFB3104703, NSFC through grant 62002007, Natural Science Foundation of Beijing Municipality through grant 4222030,  S\&T Program of Hebei through grant 21340301D, the Fundamental Research Funds for the Central Universities, and Xiaomi Young Scholar Funds for Beihang University.
Philip S. Yu was supported by NSF under grants III-1763325, III-1909323, III-2106758, and SaTC-1930941.
For any correspondence, please refer to Hao Peng.
\end{acks}

\bibliographystyle{ACM-Reference-Format}
\balance
\bibliography{sample-base}

\appendix

\section{Data Statistics}
We present the detailed datasets statistics in Table \ref{tab:data_stat}. We evaluate all models in four benchmark datasets from the public Amazon review dataset\footnote{\url{http://deepyeti.ucsd.edu/jianmo/amazon/index.html}}. In the Amazon reviews dataset, there are multiple categories of product interactions with timestamps from users. We choose \textit{Beauty}, \textit{Toys and Games}~(Toys), \textit{Tools and Home}~(Tools), and \textit{Office Products}~(Office) categories in our experiments as these four categories are widely used benchmark datasets~\cite{tang2018personalized, kang2018self, sun2019bert4rec, fan2021modeling, li2020time, fan2022sequential}. We treat the presence of user-item reviews as user-item interactions. For each user, we sort the interacted items based on the timestamp to form the interaction sequence. In each user sequence, we use the last interaction for testing and the second to last one for validation. We adopt the standard 5-core pre-processing step on users~\cite{tang2018personalized, kang2018self, sun2019bert4rec, fan2021modeling, li2020time, fan2022sequential} to filter out users with less than five interactions. We present detailed datasets statistics in Table~\ref{tab:data_stat}.
\begin{table}[H]
\centering
\caption{Datasets Statistics.}
\label{tab:data_stat}
\resizebox{0.48\textwidth}{!}{%
\begin{tabular}{@{}crrrrc@{}}
\toprule
Dataset & \multicolumn{1}{c}{\#users} & \multicolumn{1}{c}{\#items} & \multicolumn{1}{c}{\#interactions} & \multicolumn{1}{c}{density} & \begin{tabular}[c]{@{}c@{}}avg. \\ interactions \\ per user\end{tabular} \\ \midrule
Beauty & 22,363 & 12,101 & 198,502 & 0.05\% & 8.3 \\
Toys & 19,412 & 11,924 & 167,597 & 0.07\% & 8.6 \\
Tools & 16,638 & 10,217 & 134,476 & 0.08\% & 8.1 \\
Office & 4,905 & 2,420 & 53,258 & 0.44\% & 10.8 \\
\bottomrule
\end{tabular}%
}
\end{table}

\section{Evaluation}
We generate the top-N recommendation list for each user based on the sequence-item Wasserstein distance in ascending order. We \textbf{rank all items} for all models so that no sampling bias is introduced in evaluation~\cite{krichene2020sampled}. The evaluation includes standard top-N ranking metrics, Recall@N, NDCG@N, and MRR. 
We report the average results over all test users. The test results are reported based on the best validation results. We report metrics in multiple Ns, including $N=\{1, 5, 10\}$, which are widely adopted in~\cite{kang2018self, sun2019bert4rec, fan2022sequential}.

\section{Hyper-parameters Grid Search}
We implement \modelname with Pytorch. We grid search all parameters and report the test performance based on the best validation results.
For all baselines, we search the embedding dimension in $\{64, 128\}$. 
As the proposed model has both mean and covariance embeddings, we only search for $\{32, 64\}$ for \modelname for the fair comparison. 
We also search max sequence length from $\{50, 100\}$.  
We tune the learning rate in $\{10^{-3},10^{-4}\}$, search the $L2$ regularization weight from $\{10^{-1}, 10^{-2}, 10^{-3}\}$, dropout rate from $\{0.3, 0.5, 0.7\}$. 
For sequential methods, we search number of layers from $\{1,2,3\}$, and number of heads in $\{1,2,4\}$. 
We adopt the early stopping strategy that model optimization stops when the validation MRR does not increase for 50 epochs. 
The followings are the model specific hyper-parameters search ranges of baselines:
\begin{itemize}[leftmargin=*]
    \item \textbf{BPR\footnote{\url{https://github.com/xiangwang1223/neural_graph_collaborative_filtering}}:} BPR is the most classical collaborative filtering method for personalized ranking with implicit feedbacks. We search the learning rate in $\{10^{-3},10^{-4}\}$ and $L2$ regularization weight from $\{10^{-1}, 10^{-2}, 10^{-3}\}$.
    \item \textbf{Caser\footnote{\url{https://github.com/graytowne/caser_pytorch}}:} A CNN-based sequential recommendation method that views the sequence embedding matrix as an image and applies convolution operators to it. We search the length $L$ from $\{5, 10\}$, and $T$ from $\{1, 3, 5\}$.
    \item \textbf{SASRec\footnote{\url{https://github.com/RUCAIBox/CIKM2020-S3Rec}}:} The state-of-the-art sequential method that depends on the Transformer architecture. We search the dropout rate from $\{0.3, 0.5, 0.7\}$. 
    \item \textbf{BERT4Rec\footnote{\url{https://github.com/FeiSun/BERT4Rec}}:} This method extends SASRec to model bidirectional item transitions with standard Cloze objective. We search the mask probability from the range of $\{0.1, 0.2, 0.3, 0.5, 0.7\}$.
    \item \textbf{STOSA\footnote{\url{https://github.com/zfan20/STOSA}}:} A metric learning-base sequential method that models items as distributions and proposes a Wasserstein self-attention module. We search the dropout rate from $\{0.3, 0.5, 0.7\}$.
    \item \textbf{CL4Rec:}\footnote{\url{https://github.com/YChen1993/CoSeRec}} A sequential recommendation method that introduces masking, reorder, and cropping data augmentations in the contrastive learning framework. We search the masking rate from $\{0.1, 0.2, 0.3, 0.4, 0.5\}$ and cropping ratio from $\{0.1, 0.2, 0.3, 0.4, 0.5\}$.
    \item \textbf{DuoRec:}\footnote{\url{https://github.com/RuihongQiu/DuoRec}} This method introduces unsupervised Dropout and supervised semantic augmentations in self-supervised learning for sequential recommendation. 
    \item \textbf{CoSeRec:}\footnote{\url{https://github.com/YChen1993/CoSeRec}} This method extends CL4Rec with additional data augmentation techniques.
\end{itemize}

\section{Users and Items Improvements Analysis on All Datasets}
Detailed analysis and observations can be found in Section~\ref{sec:imp_analys}.
\begin{figure}
\begin{subfigure}[t]{0.235\textwidth}
    \
    \includegraphics[width=\textwidth]{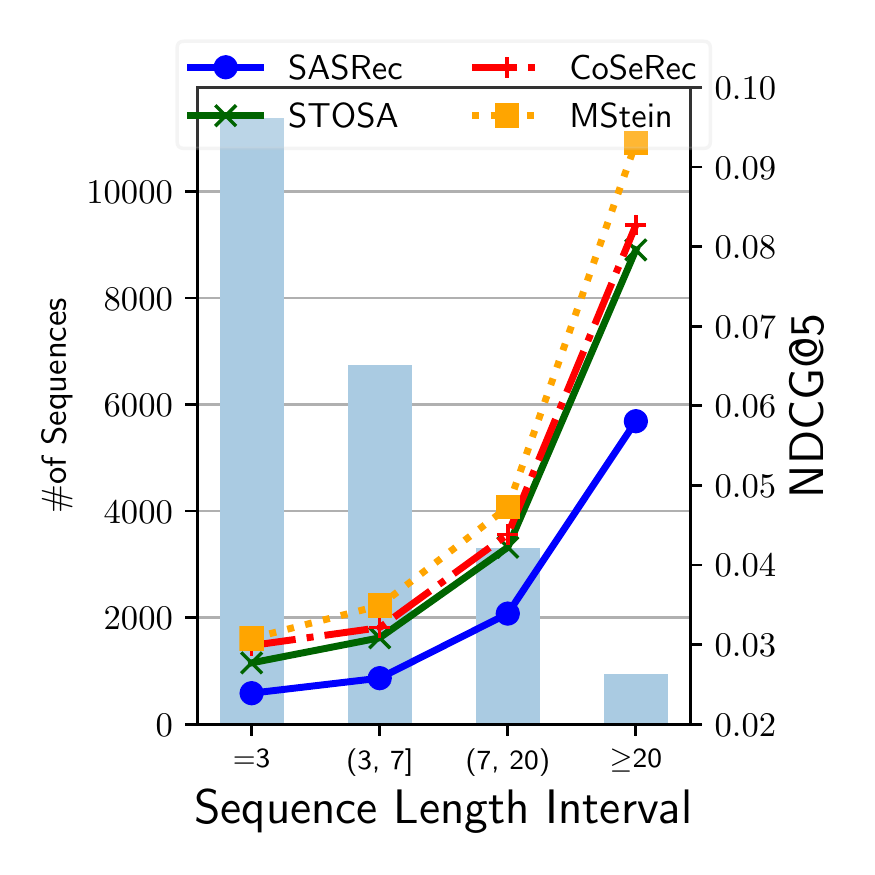}
    \caption{Beauty}
    \label{fig:beauty_item}
\end{subfigure}
\begin{subfigure}[t]{.235\textwidth}
    \includegraphics[width=\textwidth]{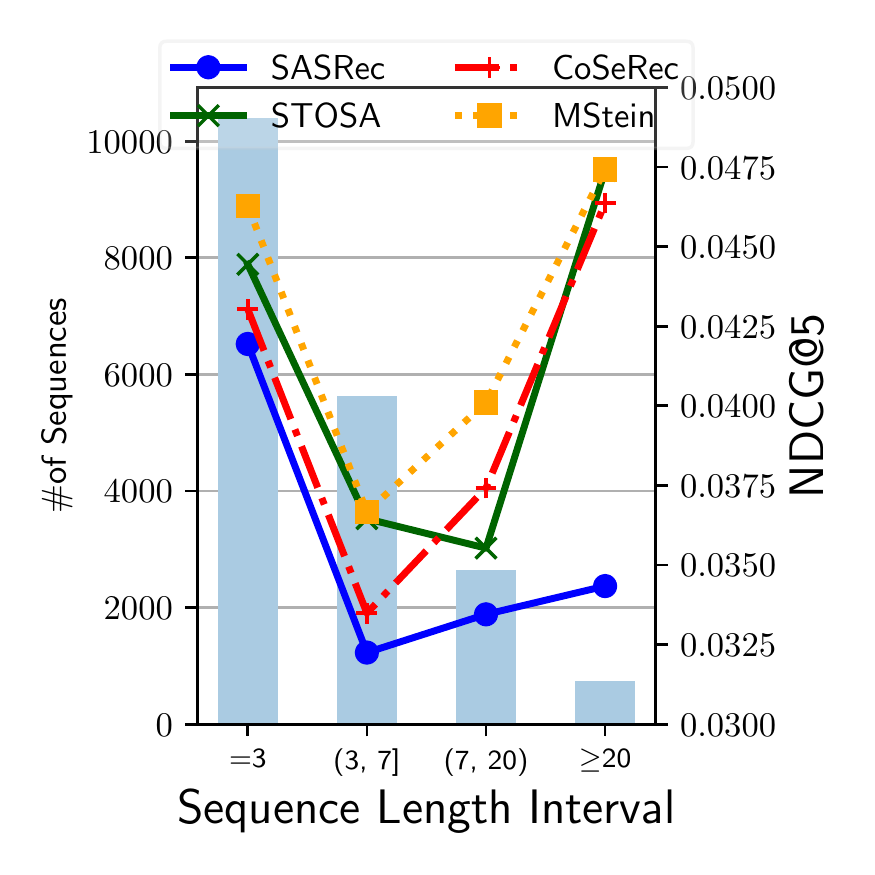}
    \caption{Toys}
    \label{fig:toys_item}
\end{subfigure}
\\
\begin{subfigure}[t]{0.235\textwidth}
    \
    \includegraphics[width=\textwidth]{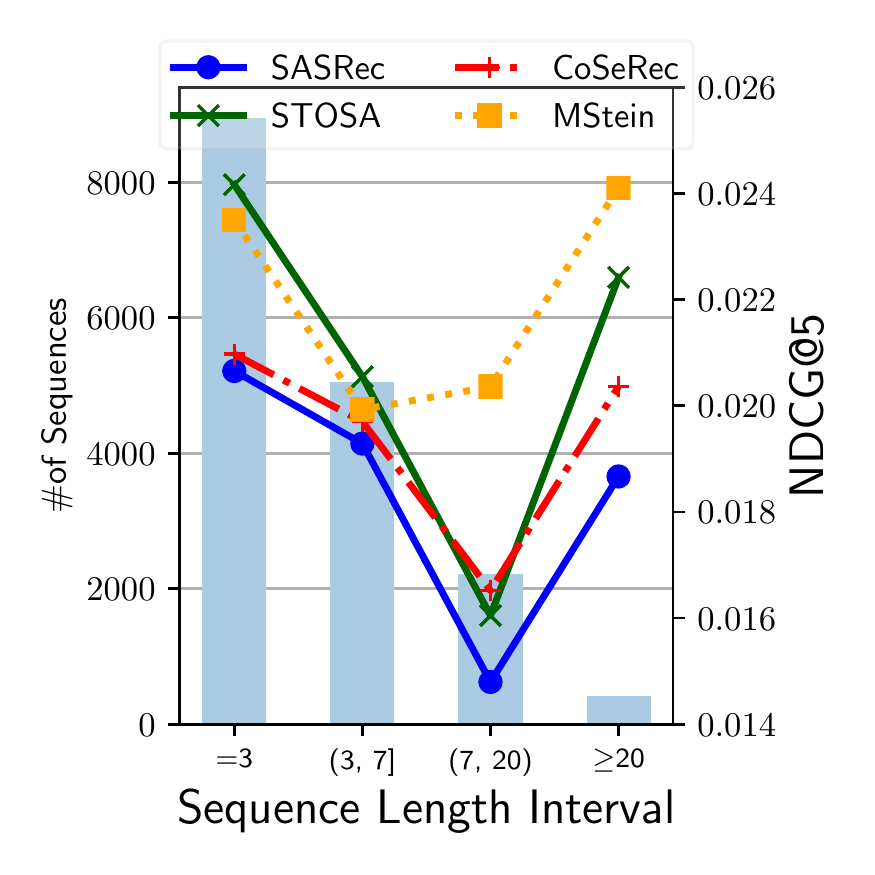}
    \caption{Tools}
    \label{fig:tools_item}
\end{subfigure}
\begin{subfigure}[t]{.235\textwidth}
    \includegraphics[width=\textwidth]{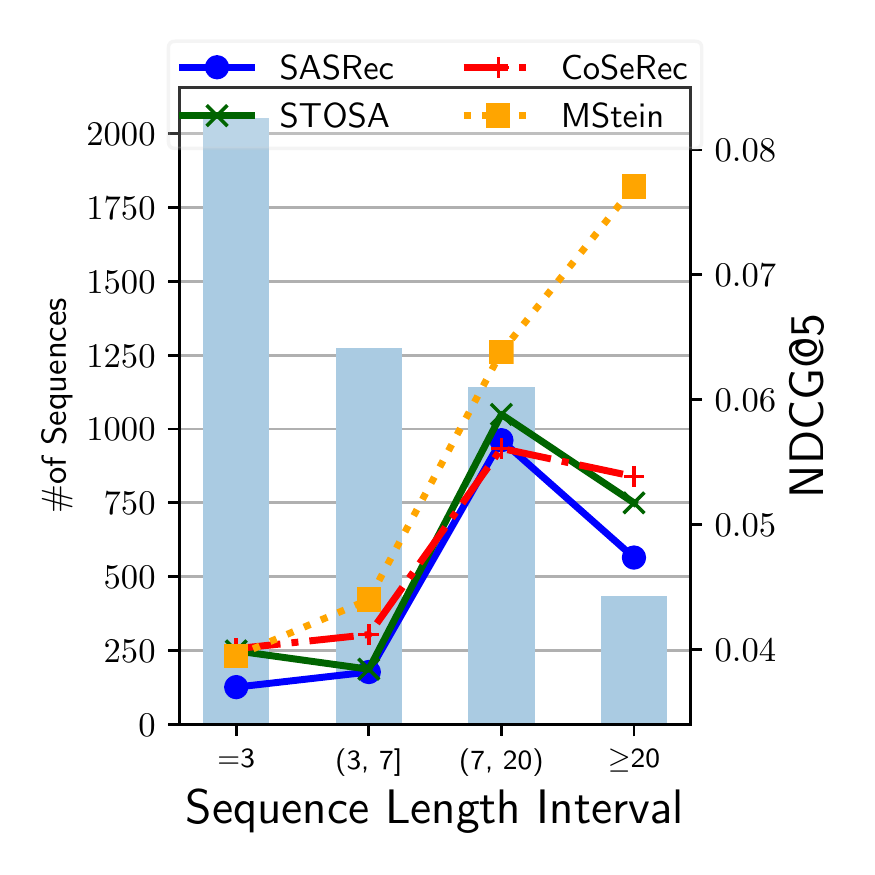}
    \caption{Office}
    \label{fig:office_item}
\end{subfigure}
\vspace{-3mm}
\caption{NDCG@5 on different sequences based on length.}
\label{fig:ndcg5_seqlen}
\end{figure}
\begin{figure}
\begin{subfigure}[t]{0.235\textwidth}
    \
    \includegraphics[width=\textwidth]{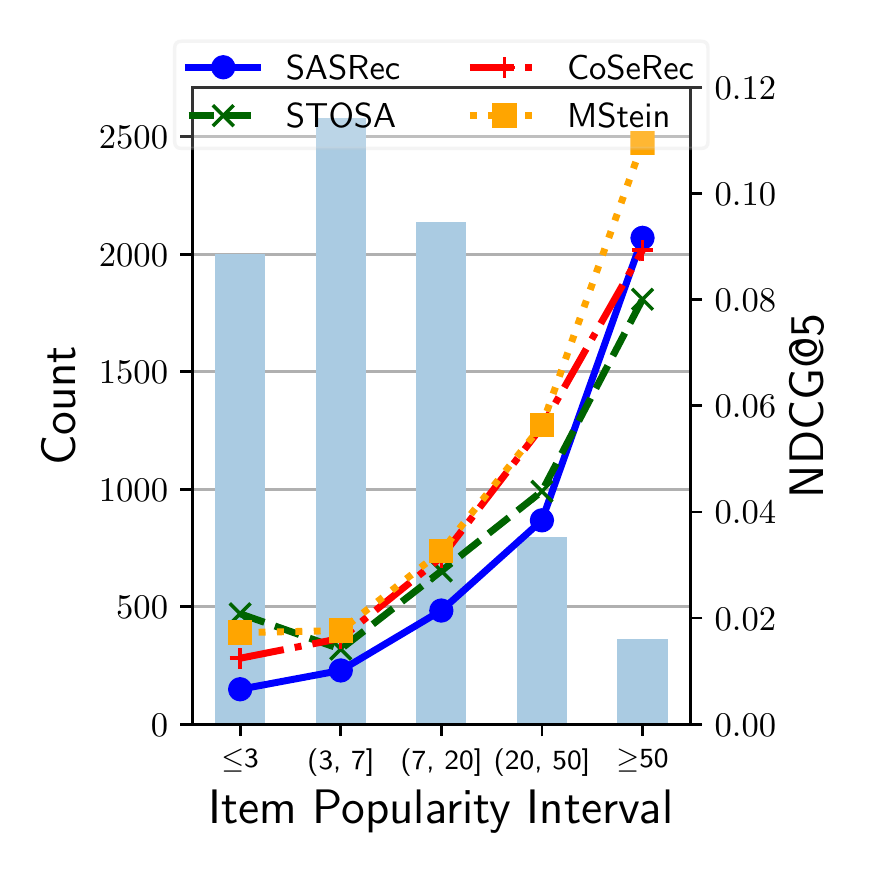}
    \caption{Beauty}
    \label{fig:beauty_item}
\end{subfigure}
\begin{subfigure}[t]{.235\textwidth}
    \includegraphics[width=\textwidth]{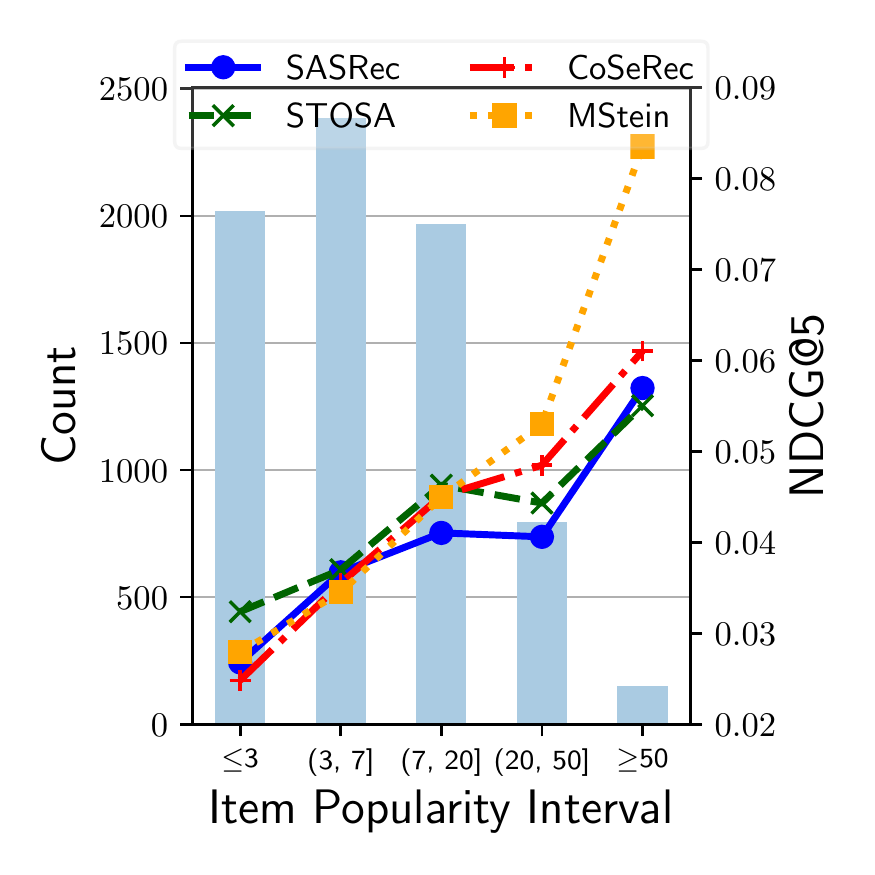}
    \caption{Toys}
    \label{fig:toys_item}
\end{subfigure}
\\
\begin{subfigure}[t]{0.235\textwidth}
    \
    \includegraphics[width=\textwidth]{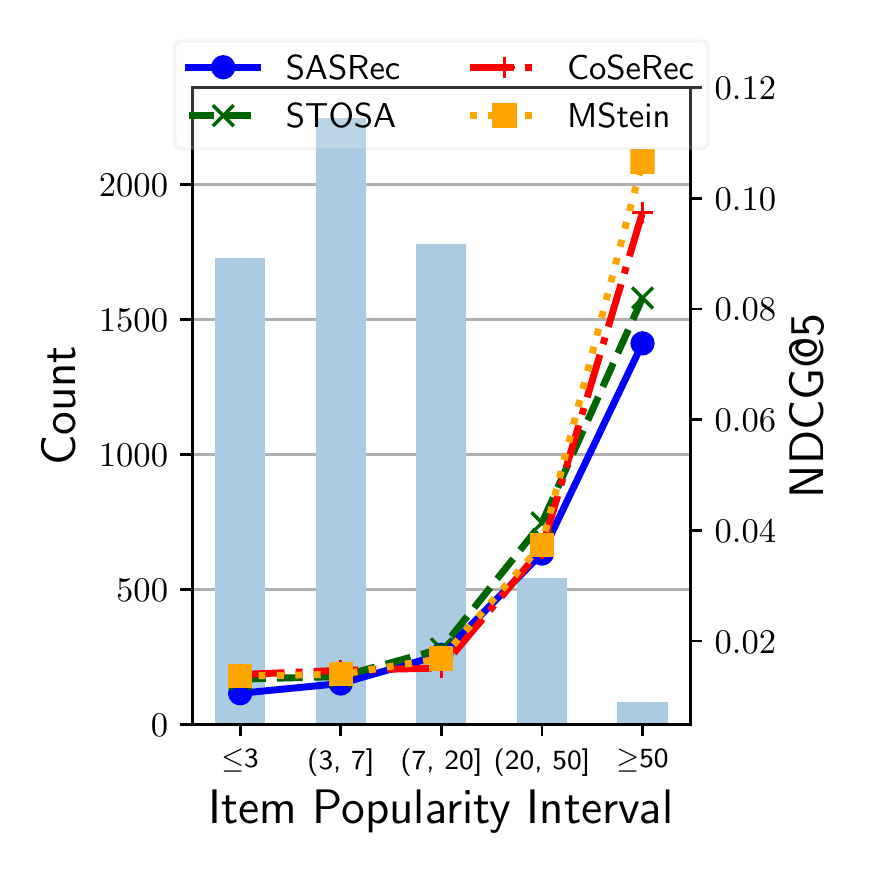}
    \caption{Tools}
    \label{fig:tools_item}
\end{subfigure}
\begin{subfigure}[t]{.235\textwidth}
    \includegraphics[width=\textwidth]{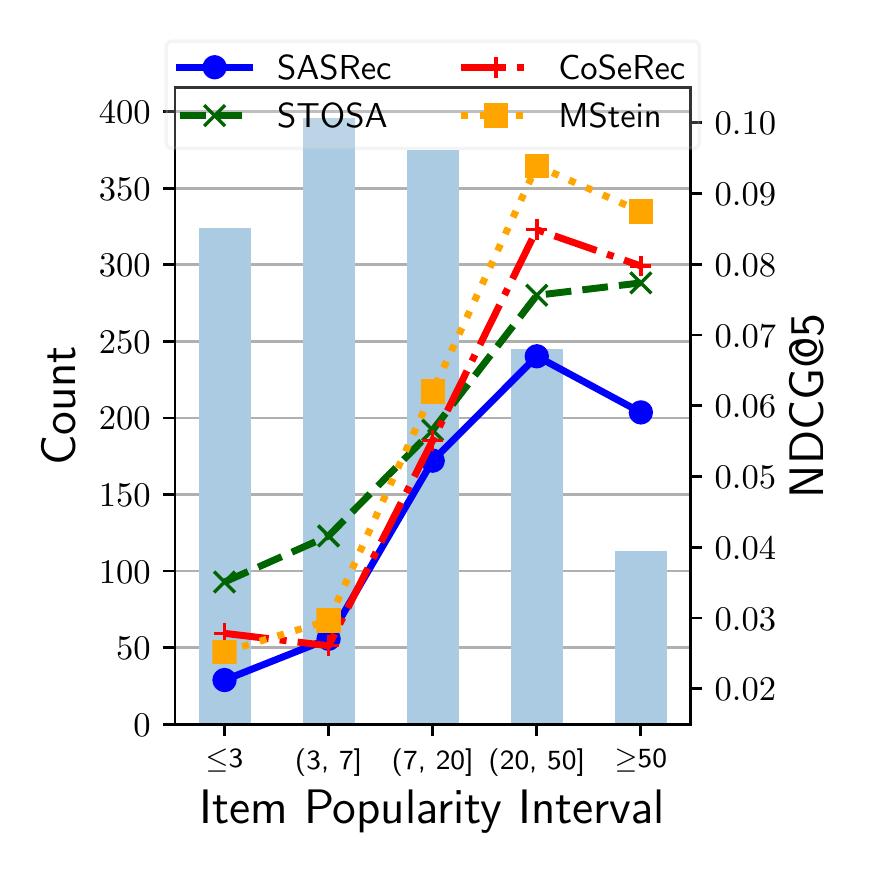}
    \caption{Office}
    \label{fig:office_item}
\end{subfigure}
\caption{NDCG@5 on different items based on popularity.}
\label{fig:item_pop_full}
\end{figure}

\end{document}